\documentclass[11pt]{article}

\usepackage[margin=1in]{geometry}
\usepackage[T1]{fontenc}
\usepackage{lmodern}
\usepackage{microtype}
\usepackage[numbers,sort&compress]{natbib}
\usepackage[hidelinks]{hyperref}
\usepackage{graphicx}
\usepackage{subfigure}
\usepackage{amsthm}
\usepackage[ruled,linesnumbered,lined]{algorithm2e}
\SetKwInOut{Input}{input}
\SetKwInOut{Output}{output}

\usepackage{amsmath,amsfonts,bm}

\def\eqref#1{equation~\ref{#1}}

\def\1{\bm{1}}

\def\vm{{\bm{m}}}

\def\vw{{\bm{w}}}

\DeclareMathAlphabet{\mathsfit}{\encodingdefault}{\sfdefault}{m}{sl}
\SetMathAlphabet{\mathsfit}{bold}{\encodingdefault}{\sfdefault}{bx}{n}

\newcommand{\E}{\mathbb{E}}

\newcommand{\R}{\mathbb{R}}

\DeclareMathOperator*{\argmin}{arg\,min}

\theoremstyle{plain}
\newtheorem{theorem}{Theorem}[section]
\newtheorem{lemma}[theorem]{Lemma}
\newtheorem{proposition}[theorem]{Proposition}
\newtheorem{corollary}[theorem]{Corollary}

\theoremstyle{definition}
\newtheorem{definition}[theorem]{Definition}
\newtheorem{remark}[theorem]{Remark}

\renewcommand{\eqref}[1]{\textup{(\ref{#1})}}

\newcommand{\C}{{\mathbb C}}
\newcommand{\diag}[1]{\textbf{diag}\left(#1\right)}

\newcommand{\norm}[1]{\left\lVert#1\right\rVert}
\newcommand{\abs}[1]{\left\lvert#1\right\lvert}

\hypersetup{
  pdftitle={Model Selection and Parameter Estimation of One-Dimensional Gaussian Mixture Models},
  pdfauthor={Xinyu Liu and Hai Zhang},
  pdfsubject={Gaussian mixture model selection and parameter estimation},
  pdfkeywords={Gaussian mixture models, model-order selection, minimax testing, Fourier methods, sample complexity}
}
\title{Model Selection and Parameter Estimation of One-Dimensional Gaussian Mixture Models}
\author{
Xinyu Liu$^{1}$ \quad Hai Zhang$^{1,2}$\\[0.5em]
\small $^{1}$Department of Mathematics, The Hong Kong University of Science and Technology\\
\small $^{2}$HKUST--Shenzhen--Hong Kong Collaborative Innovation Research Institute\\[0.3em]
\small \texttt{xliuem@connect.ust.hk} \quad \texttt{haizhang@ust.hk}
}
\date{}

\begin{document}
\maketitle

\begin{abstract}
We study exact model-order identification for one-dimensional Gaussian location
mixtures with a known common variance in the poorly separated regime. We
formulate binary testing between $k$- and $(k-1)$-component mixtures under
bounded-support, minimum-weight, and minimum-separation conditions. A
moment-matched construction and Le~Cam's two-point method show that reliable
identification in the fully clustered worst-case configuration requires
$\Omega(\Delta^{-(4k-4)})$ samples. We propose an oracle Fourier thresholding
procedure based on the singular values of a Hankel matrix formed from
deconvolved empirical characteristic functions. Under the stated conditions,
it recovers the model order with probability at least $1-\delta$ using
$O(w_{\min}^{-2}\Delta^{-(4k-4)}\log(1/\delta))$ samples. Hence, for fixed
remaining parameters, its separation exponent matches the minimax lower bound.
When only one pair of components collides, the upper bound improves to
$O(\Delta^{-4})$ and matches the corresponding two-point lower-bound exponent;
more generally, a merging cluster of $m$ components yields the sufficient bound
$O(\Delta^{-4(m-1)})$. We also develop a MUSIC-based extension for estimating
the mixing distribution and present numerical experiments.
\end{abstract}

\medskip
\noindent\textbf{Keywords:} Gaussian mixture models; model-order selection;
minimax testing; Fourier methods; sample complexity.

\section{Introduction}
Mixture models are widely employed across various fields to model data and signals originating from sub-populations or distinct sources. Among these models, the Gaussian mixture model (GMM) has emerged as one of the most extensively studied and widely applied models. It has found broad applications in machine learning, pattern recognition, image processing, etc. A GMM represents a probability distribution as a convex combination of several Gaussian components, each of which is characterized by its own mean and covariance matrix. Formally, each observation of the GMM follows:
    \[
        X \sim \sum_{i=1}^k w_i \mathcal{N}(\mu_i, \Sigma_i),
    \]
where $w_i$ is the mixing weight such that $w_i > 0$ and $\sum_{i=1}^k w_i = 1$ and $\mu_i, \Sigma_i$ are the mean and covariance matrix of the $i$-th component, respectively. We can introduce a latent variable $Z\in \{1, \cdots , k\}$, with marginal distribution specified by the mixing weights $\mathbb{P}(Z = i) = w_i$.
Thus, the GMM can also be expressed conditionally as 
    \begin{equation}
        \label{eqn:conditional}
        X \mid (Z=i) \sim \mathcal{N}(\mu_i, \Sigma_i).
    \end{equation}
Given samples drawn from the distribution, the central challenge is to learn the underlying mixture model. Generally, there are three formulations for learning mixtures: 
\begin{itemize}
    \item \textit{Clustering}: estimate the latent variable $z_j$ for each sample $x_j$;
    \item \textit{Parameter estimation}: estimate the weights $w_i$'s, means $\mu_i$'s and covariance matrix $\Sigma_i$'s up to a global permutation;
    \item \textit{Density estimation}: estimate the probability density function of the GMM under a specific loss function.
\end{itemize}  

Given the extensive literature surrounding these three learning problems, we offer a brief review of the most relevant methods, emphasizing the intricate difficulty of the learning task in relation to the separation or statistical differences between components.  The most common clustering approach is $k$-means, which aims to minimize:
    \begin{equation}
        \label{eqn:kmeans}
        \argmin_{z_j, \mu_i}\sum_{j=1}^n \sum_{i=1}^k \mathbf{1}\left\{z_j = i\right\} \norm{x_j - \mu_i}^2,
    \end{equation}
where $\mathbf{1}\left\{z_j = i\right\} = 1$ if $z_j = i$ otherwise 0. It is well-known that solving the $k$-means exactly in the general case is NP-hard, even for two clusters (see \cite{aloise2009np}). Various computationally tractable approximation approaches have been proposed, including the widely used Lloyd's algorithm (\cite{lloyd1982least}), nonnegative matrix factorization (NMF) (see \cite{paatero1994positive, he2011symmetric, zhuang2023statistically}), and semidefinite programming (SDP) (see \cite{peng2007approximating}). Note that Lloyd's algorithm iterates a two-phase of re-assigning the samples to clusters and re-computing the cluster means until convergence. The perfect clustering of the mixture depends on the separation distance defined as $\Delta := \min_{1\leq i < j \leq k} \norm{\mu_i - \mu_j}$.
It has been shown in \cite{ndaoud2022sharp} that the critical threshold for a perfect clustering of a two-component Gaussian mixture with a unified covariance matrix $\sigma^2 I$ in $d$-dimension is: 
    \begin{equation}
        \label{eqn:perfect clustering 2}
        {\Delta}^2 = \sigma^2 \left(1 + \sqrt{1 + \frac{2d}{n\log n}}\right)\log n.
    \end{equation}
Similar results are obtained for the $k$-component mixture model in \cite{chen2021cutoff}.  

Parameter estimation and density estimation benefit from a larger sample size, contrasting with the perfect clustering scenario in \eqref{eqn:perfect clustering 2}. Existing methodologies for learning the mixture can be broadly categorized into the maximum-likelihood method and the moment-based method. The maximum likelihood method aims to maximize the likelihood of the given samples. 
Numerous iterative methods for optimization are proposed to seek the maximum or local maximum of the likelihood function. Among them, the most widely used one is the EM (Expectation-Maximization) Algorithm (\cite{dempster1977maximum}). The EM algorithm iterates a two-step operation to find a local maximum of the log-likelihood function, which may not necessarily be the ground-truth parameters. 
The moment-based methods date back to \cite{pearson1894contributions}. However, Pearson’s method has practical limitations due to its sensitivity to moment selection and the instability of finding roots of high-degree polynomials. Various modifications of the method of moments are proposed, such as the Generalized Method of Moments(\cite{hansen1982large}) and the Denoised Method of Moments(\cite{wu2020optimal}), where the latter established the optimal convergence rate for estimating the mixing distribution under the Wasserstein distance and was extended into high-dimensional GMMs in \cite{doss2023optimal}.

In addition, \cite{vempala2002spectral} proposed a spectral projection method for learning mixtures of spherical Gaussians, subject to a separation condition based on component variances. \cite{hsu2013learning} offered an alternative technique utilizing the spectral decomposition of low-order observable moments, under an assumption on the linear independence of the means. \cite{moitra2010settling} established a polynomial-time, polynomial-sample algorithm for general multi-dimensional GMMs via systems of moments and projections, where the statistical difference of components strictly binds the sample complexity. More recently, \cite{regev2017learning} identified a sharp threshold for the sample complexity of $k$ standard spherical GMMs: learning requires super-polynomial samples if the separation is $o(\sqrt{\log k})$, whereas polynomial samples suffice if the separation is $\Omega(\sqrt{\log k})$.


It is important to note that standard clustering algorithms such as
$k$-means, as well as parameter estimation methods based on maximum
likelihood or moments, typically assume that the model order is known in
advance. Misspecifying the model order can substantially affect subsequent
parameter estimation. Let $k^\star$ denote the true number of components
and let $m$ denote the fitted model order. When the model is underspecified,
i.e., $m<k^\star$, Leroux \cite{leroux1992consistent} showed that, under
suitable regularity conditions, maximum-likelihood estimators constrained
to the $m$-component model converge weakly to the class of $m$-component
mixing distributions that minimize the Kullback--Leibler divergence from
the true mixture. Consequently, underfitting generally introduces an
irreducible approximation bias and prevents recovery of the true latent
structure. When the model is overspecified, i.e., $m>k^\star$, the true mixing
distribution lies in a singular part of the fitted parameter space, since
redundant components may have vanishing weights or may merge with existing
components. \cite{chen1995optimal} first demonstrated that the optimal convergence rate for estimating the mixing distribution deteriorates from $\mathcal{O}(n^{-1/2})$ when the model order is overspecified. The initially conjectured local rate of $\mathcal{O}(n^{-1/4})$ was later refined by \cite{heinrich2018strong}. They showed that, under
strong identifiability conditions, the optimal local minimax rate for
estimating a mixing distribution with at most $m$ components around a
$k^\star$-component mixing distribution is
\[
    n^{-1/(4(m-k^\star)+2)}
\]
in the Wasserstein distance. This rate reduces to the parametric rate
$n^{-1/2}$ when $m=k^\star$, but becomes progressively slower as more
redundant components are allowed. However, since $k$ is frequently unknown a priori, effective methods for determining the appropriate model order are essential. To address this, various statistical and information-theoretic criteria have been proposed, such as the Akaike Information Criterion (AIC) and the Bayesian Information Criterion (BIC). These measures quantify the trade-off between model complexity and data fidelity (see \cite{konishi2008information}).

Specific advancements include \cite{maugis2012non}, which offers a non-asymptotic penalized criterion for GMMs. Alternatively, \cite{corduneanu2001variational} employ variational inference, selecting the model order by placing prior distributions on parameters and maximizing the posterior. Other common techniques include cross-validation and likelihood-ratio tests (\cite{mclachlan2014number}). Despite this extensive literature, rigorous theoretical guarantees for the model selection problem remain scarce. The primary objective of this paper is to identify an explicit difficult family of closely spaced mixtures and quantify the sample size required to distinguish each such mixture from a moment-matched lower-order approximation under a likelihood-based criterion. Furthermore, we propose an efficient algorithm utilizing Fourier data, derive a sufficient recovery guarantee, and compare its separation-dependent scaling with that of the hard-instance analysis.

\subsection{Problem Setting and Main Contributions}
In this paper, we restrict to one-dimensional Gaussian mixture models (GMMs) with a known unified variance. Specifically, 
for $\sigma>0$, let
\[
    \varphi_\sigma(x)
    := \frac{1}{\sqrt{2\pi}\sigma}
       \exp\left(-\frac{x^2}{2\sigma^2}\right)
\]
denote the density of $\mathcal{N}(0,\sigma^2)$. Consider the probability distribution
    \begin{equation}
    \label{eqn:model}
         P = \sum_{i=1}^k w_i\mathcal{N}(\mu_i, \sigma^2),
         \qquad \sum_{i=1}^k w_i = 1,
    \end{equation}
with density
\[
    p(x) = \sum_{i=1}^k w_i\varphi_\sigma(x-\mu_i).
\]
Here $\{\mu_i\}_{i=1}^k$ and $\{w_i\}_{i=1}^k$ are the component means and weights, respectively, and all components have the common variance $\sigma^2$.
We refer to $k$ as the model order. We assume that the means are bounded by $|\mu_i| < R$, $1\leq i\leq k$ for some $R>0$. This model is also known as the Gaussian location mixture. Equivalently, we can express the model as 
    \begin{equation}
        P = \nu_P * \mathcal{N}(0, \sigma^2),
        \qquad
        p(x) = \int_{\mathbb{R}} \varphi_\sigma(x-u)\,\mathrm{d}\nu_P(u),
    \end{equation}
 where $\nu_P = \sum_{i=1}^k w_i \delta_{\mu_i}$ represents the mixing distribution. We define the separation distance $\Delta$ and the minimal weight $w_{\min}$ of the model as 
\begin{equation} \label{eq-sep}
     \Delta = \min_{1\leq i < j \leq k} \abs{\mu_i - \mu_j}, \quad w_{\min} = \min_{1\leq i \leq k}w_i.
\end{equation}

Given independent and identically distributed samples
$X_1,\ldots,X_n\stackrel{\mathrm{i.i.d.}}{\sim}P$, our goal is to estimate the mixture model parameters and select the model order $k$.

We refer the reader to Hardt and Price \cite{hardt2015tight} for parameter estimation (excluding model order selection) in general one-dimensional two-Gaussian mixtures. They show that achieving $\varepsilon$-accurate estimation with probability at least $1 - \delta$ requires $O\!\left(\varepsilon^{-12} \log \tfrac{1}{\delta}\right)$ samples in the worst case. However, when the two components are separated by a distance on the order of the mixture’s standard deviation, the sample complexity improves to $O\!\left(\varepsilon^{-2} \log \tfrac{1}{\delta}\right)$.

We focus on the ``super-resolution'' regime, where the component means are poorly separated, i.e., the separation $\Delta$ is much smaller than the variance $\sigma$ so that samples from different components overlap with high probability. Here, we draw an analogy to super-resolution in optical systems, where the objective is to distinguish light sources separated below the diffraction limit using their optical measurements. In this challenging setting, we address two questions: (i) How does the sample size needed by our likelihood-based distinguishability criterion scale for an explicit difficult family? and (ii) Can a computationally efficient Fourier-based procedure attain the same separation exponent under suitable assumptions? Despite the vast literature on GMM estimation, these questions have rarely been examined in depth.

The contributions of the paper can be summarized as follows:
\begin{itemize}
    \item We formulate model-order identification as a binary testing problem between $k$- and $(k-1)$-component Gaussian location mixtures over explicitly specified parameter classes. Using a moment-matched construction together with Le~Cam's two-point method, we establish a procedure-independent minimax lower bound showing that, in the fully clustered worst-case regime, reliable model-order identification requires
    \[
        n=\Omega\!\left(\Delta^{-(4k-4)}\right)
    \]
    for fixed model order, mixing weights, and common variance. We further show that the difficulty depends on the local collision geometry: when only one pair of components approaches collision while the remaining components stay resolved, the corresponding two-point lower-bound exponent improves to $4$.

    \item We propose an oracle Fourier thresholding procedure based on the singular values of a Hankel matrix constructed from deconvolved empirical characteristic-function measurements. Under the stated bounded-support, minimum-weight, separation, and threshold assumptions, the procedure controls both testing errors and recovers the correct model order with probability at least $1-\delta$ using
    \[
        n=O\!\left(
            w_{\min}^{-2}\Delta^{-(4k-4)}
            \log\frac{1}{\delta}
        \right)
    \]
    samples in the fully clustered regime. When only one pair of components collides, the sufficient dependence improves to $O(\Delta^{-4})$, matching the corresponding two-point lower-bound exponent. Thus, for fixed remaining parameters, the oracle Fourier procedure achieves the minimax-optimal separation exponent in both the fully clustered and single-collision regimes. More generally, when $m$ components form a merging cluster, our analysis gives the sufficient dependence $O(\Delta^{-4(m-1)})$. We also develop a MUSIC-based extension for mixing-distribution estimation and evaluate the proposed methods numerically.
\end{itemize}

The main distinction from \cite{heinrich2018strong} and
\cite{wu2020optimal} is the statistical target. Both works study estimation of
the complete mixing distribution under Wasserstein loss: Heinrich and Kahn give
local minimax rates around a fixed lower-order mixing distribution, whereas Wu
and Yang establish worst-case and adaptive rates for Gaussian location mixtures.
Their main results do not formulate exact identification of the model order as a
$k$-versus-$(k-1)$ testing problem. In contrast, our risk is a testing error and
our objective is to determine whether an additional component is present under
explicit minimum-separation and minimum-weight conditions. Our constructive
upper bound also uses a different mechanism---Fourier deconvolution followed by
Hankel-matrix rank estimation---rather than minimum-distance or denoised-moment
estimation.

This distinction explains why the rates are not directly comparable. In our
order-testing construction, the $k$- and $(k-1)$-point mixing distributions match
moments through order $2k-3$, so the first unmatched moment has order $2k-2$ and
leads to the separation exponent $4k-4$. In the least-favorable constructions for
estimating a general $k$-atomic mixing distribution, moments can match through
order $2k-2$, leading instead to the Wasserstein estimation exponent $4k-2$.
Thus, the difference between the exponents reflects different losses and
parameter classes and should not be interpreted as saying that model-order
testing is uniformly easier than mixing-distribution estimation.

\subsection{Paper Organization and Notations}
The rest of the paper is organized as follows. In Section \ref{sec:crl}, we formulate a likelihood-based criterion, analyze an explicit hard family, and establish minimax lower bounds by Le~Cam's method. In Section \ref{sec:fourier}, we propose several algorithms based on Fourier measurements from finite samples and derive a sufficient recovery guarantee. We then compare the separation exponent in this guarantee with that arising from the lower-bound analyses. In Section \ref{sec:numerical}, we conduct several numerical experiments to compare the accuracy and efficiency of the proposed algorithms with commonly used algorithms for both model selection and mixing distribution estimation.

Throughout the paper, we write $f(x) = O(g(x))$ if there exists some constant $c_1>0$ such that $f(x) < c_1 g(x)$, and $f(x) = o(g(x)),\ x\to 0$ if $\lim_{x\to 0}{f(x)}/{g(x)} = 0$. Define the cumulative distribution function of standard normal as 
    \[
        \Phi(x) = \mathbb P_{Z\sim \mathcal{N}(0,1)}(Z \leq x) = \frac{1}{\sqrt{2\pi}}\int_{-\infty}^x e^{-\frac{u^2}{2}}\,du.
    \]
We shall use the following constant
    \begin{equation}
    \label{eqn:zeta}
        \zeta(k) = 
        \begin{cases} 
        \left( \frac{(k-1)}{2}! \right)^2, & \text{if } k \text{ is odd}, \\ 
        \left( \frac{k}{2}! \right) \left( \frac{k-2}{2}! \right), & \text{if } k \text{ is even}.
        \end{cases}
    \end{equation}

\section{Lower Bounds of Sample Complexity for Model Order Selection}
\label{sec:crl}
In this section, we identify a difficult regime for likelihood-based model-order comparison. Given a $k$-component model $P$ with density $p$, a lower-order model $Q$ with density $q$, and samples $X_1,\ldots,X_n\stackrel{\mathrm{i.i.d.}}{\sim}P$, the likelihood-based criterion favors $P$ over $Q$ when
\[
    \frac{1}{n}\sum_{j=1}^n\log p(X_j)
    >
    \frac{1}{n}\sum_{j=1}^n\log q(X_j).
\]
We first construct explicit GMM families for which this criterion fails to favor the true, higher-order model with nonnegligible probability at finite sample sizes. These are hard-instance lower bounds for the specified criterion. We then use Le~Cam's method to derive unconditional minimax lower bounds over the parameter classes stated below.

\subsection{Likelihood-based Sample Complexity for Two-component GMMs} \label{sec-likeli-sample}
We first consider the following two-component Gaussian mixture distribution:
\begin{equation}
    \label{eqn:two-component GMM}
    P=w\mathcal{N}(-\mu, \sigma^2) + (1-w) \mathcal{N}(\mu, \sigma^2),
\end{equation}
whose density is
\[
    p(x)=w\varphi_\sigma(x+\mu)+(1-w)\varphi_\sigma(x-\mu),
\]
where $\mu > 0$ and $0 < w \leq \frac{1}{2}$.
Let $X_1,\ldots,X_n\stackrel{\mathrm{i.i.d.}}{\sim}P$. The aim of this section is to determine the sample complexity required for estimating the model order $k=2$.
For each $m\in\mathbb{R}$, let
\[
    Q_m:=\mathcal{N}(m,\sigma^2),
    \qquad
    q_m(x):=\varphi_\sigma(x-m).
\]
Notice that we have 
    \[
        \mathbb{E}\left[
        \frac{1}{n}\sum_{j=1}^n
        \log\frac{p(X_j)}{q_m(X_j)}\right]
        = \mathbb{E}\left[\log\frac{p(X_1)}{q_m(X_1)}\right]
        = D_{\mathrm{KL}}(P\|Q_m).
    \]
In the theoretical analysis, we assume that $\mu = \frac{1}{2}\Delta \ll \sigma$ where the two components are nearly merged. In this asymptotic regime, we first construct a GMM with a single component that matches the lower order moments of $P$, as shown in the following proposition. 

\begin{proposition}
\label{prop:2-component analysis}
    Let $0<w\leq1/2$ and $\sigma>0$ be fixed, and define $r=\mu/\sigma$. Consider
    \[
        P_r=w\mathcal{N}(-r\sigma,\sigma^2)+(1-w)\mathcal{N}(r\sigma,\sigma^2),
    \]
    and let
    \[
        m_r=(1-2w)r\sigma,
        \qquad
        Q_r=\mathcal{N}(m_r,\sigma^2).
    \]
    Denote the densities of $P_r$ and $Q_r$ by $p_r$ and $q_r$, respectively. For $X\sim P_r$, define
    \[
        Y=\log\frac{p_r(X)}{q_r(X)},
        \qquad
        g(z)=2w(1-w)(z^2-1).
    \]
    Then, for $Z\sim\mathcal{N}(0,1)$, we have $\E[g(Z)]=0$ and, as $r\to0$,
    \begin{align*}
        \E[Y]&=\frac{1}{2}\E[g^2(Z)]r^4+O(r^6),\\
        \operatorname{Var}(Y)&=\E[g^2(Z)]r^4+O(r^6),\\
        \E[|Y-\E[Y]|^3]&=\E[|g(Z)|^3]r^6+o(r^6).
    \end{align*}
    In particular, for sufficiently small $r$,
    \[
        \E[|Y-\E[Y]|^3]\leq195w^3(1-w)^3r^6.
    \]
\end{proposition}
Using these moment estimations, we derive the following likelihood based sample complexity lower bound. 

\begin{theorem}
\label{thm:sample complexity 2-component}
    Under the setting of Proposition \ref{prop:2-component analysis}, for every $0<\delta<1/4$, there exist constants $r_0>0$ and $C_w>0$ such that, for every integer $n>25/\delta^2$, if
    \[
        0<r<\min\left\{r_0,\,C_w[-\Phi^{-1}(2\delta)]^{1/2}n^{-1/4}\right\},
    \]
    then the Gaussian distribution $Q_r$ constructed in Proposition \ref{prop:2-component analysis} satisfies
    \[
        \mathbb{P}\left(
        \frac{1}{n}\sum_{j=1}^n \log p_r(X_j)
        \leq
        \frac{1}{n}\sum_{j=1}^n \log q_r(X_j)
        \right) \geq \delta.
    \]
    where $X_1,\ldots,X_n\stackrel{\mathrm{i.i.d.}}{\sim}P_r$.
\end{theorem}

Theorem~\ref{thm:sample complexity 2-component} compares the average
log-likelihood of the data under the true two-component mixture $P_r$ against that
under the simpler single Gaussian $Q_r$. It asserts that, even though the
samples are genuinely drawn from $P_r$, with probability at least $\delta$ the
data are \emph{at least as consistent} with $Q_r$ as with $P_r$. In other words,
the likelihood-ratio test that prefers the mixture whenever
$$\Lambda=\tfrac1n\sum_j\log\frac{p_r(X_j)}{q_r(X_j)}>0$$ fails to do so with
probability at least $\delta$. Thus, whenever the separation $r$ is small relative
to the sample budget, the likelihood-based criterion fails to favor the true
two-component model over the single Gaussian with probability at least $\delta$.

The threshold on $r$ makes the trade-off between separation and sample size
explicit. The dominant constraint,
\[
    r \;<\; C_w\bigl[-\Phi^{-1}(2\delta)\bigr]^{1/2}\,n^{-1/4},
\]
scales as $n^{-1/4}$, so the tolerated separation shrinks only slowly as $n$
grows; equivalently, resolving a separation of size $r$ requires a sample size of
at least order $\Omega(r^{-4})$. The quartic exponent arises because $Q_r$ matches
the mean of $P_r$, so the first nonvanishing perturbation of the density is of
order $r^2$; consequently, the Kullback--Leibler divergence is of order $r^4$
(cf.~\eqref{eqn:kl-fixedvar}). It is precisely this $r^4$ divergence, accumulated
over $n$ i.i.d.\ samples, that governs the likelihood comparison. The factor
$[-\Phi^{-1}(2\delta)]^{1/2}$ reflects the failure-probability level: as
$\delta\downarrow0$, the assertion that the criterion fails with probability at
least $\delta$ becomes weaker, and the displayed sufficient condition for such a
failure allows a larger separation $r$. The auxiliary conditions $r<r_0$ and
$n>25/\delta^2$ are mild
technical requirements: the former confines $r$ to the regime where the $O(r^6)$
remainder in the KL expansion is negligible, and the latter guarantees that the
central limit approximation for the average log-likelihood ratio is accurate
enough for the stated Gaussian-quantile bound to hold.


\subsection{A Minimax lower bound of sample Complexity for two-component GMMs via Le Cam's Method}
\label{sec:minimax}

The lower bounds of sample complexity in Theorem~\ref{thm:sample complexity 2-component} concern a fixed decision rule, the sign of the
average log-likelihood ratio $\Lambda$. We now show that the same $n=\Omega(\Delta^{-4})$
scaling is in fact \emph{unconditional}: no test or estimator, however constructed, can
separate the two-component family from a single Gaussian of the same scale below this
sample size. The mechanism is Le~Cam's two-point method, with the divergence estimate established in Proposition~\ref{prop:2-component analysis}.

\paragraph{Model classes and minimax risks.}
Fix the (known) scale $\sigma>0$ and a weight $w\in(0,\tfrac12]$. For a separation
$\Delta >0$ let
\[
    \mathcal{P}_\Delta
    :=\Bigl\{\,w\,\mathcal N(\mu_1,\sigma^2)+(1-w)\,\mathcal N(\mu_2,\sigma^2)
        \;:\;\mu_1,\mu_2\in\mathbb R,\ |\mu_1-\mu_2|=\Delta\,\Bigr\}
\]
denote the order-two family with separation exactly $\Delta$, and let
\[
    \mathcal{Q}:=\bigl\{\,\mathcal N(m,\sigma^2)\;:\;m\in\mathbb R\,\bigr\}
\]
denote the order-one (null) family with the same scale. Given
$X_1,\dots,X_n\stackrel{\text{i.i.d.}}{\sim}$ (one of these laws), a \emph{test}
$\psi:\mathbb R^n\to\{0,1\}$ decides between order one ($\psi=0$) and order two
($\psi=1$). The \emph{minimax testing risk} for resolution $\Delta$ is
\begin{equation}
    \label{eqn:minimax-risk}
    \mathcal R_n(\Delta)
    :=\inf_{\psi}\Bigl[\
        \sup_{Q\in\mathcal Q}\mathbb E_{Q^{\otimes n}}[\psi]
        \;+\;
        \sup_{P\in\mathcal P_\Delta}\mathbb E_{P^{\otimes n}}[1-\psi]\
    \Bigr],
\end{equation}
the worst-case sum of Type~I and Type~II errors. A value $\mathcal R_n(\Delta)$ bounded
away from $0$ means that no test can reliably separate the two model orders at sample
size $n$. 
We have the following main result on the 
Minimax lower bound for model order. 

\begin{theorem}[Minimax lower bound for model order]
\label{thm:minimax-testing}
    Fix $w\in(0,\tfrac12]$ and $\sigma>0$, set $r=\Delta/(2\sigma)$, and let $P_r$ and $Q_r$ be defined as in Proposition~\ref{prop:2-component analysis}. For every $\varepsilon\in(0,1)$, there exists $\Delta_0>0$ such that, for every $0<\Delta\leq\Delta_0$,
    \begin{equation}
        \label{eqn:minimax-lb}
        \mathcal R_n(\Delta)
        \;\ge\;
        1-\sqrt{\tfrac{n}{2}\,D_{\mathrm{KL}}(P_r\|Q_r)}
        \;\ge\;
        1-\sqrt{1+\varepsilon}\,
        \frac{w(1-w)}{2\sqrt2}\,\frac{\Delta^2}{\sigma^2}\,\sqrt{n}.
    \end{equation}
   Consequently, for any target level $\eta\in(0,1)$, if $0<\Delta\leq\Delta_0$ and
    \[
        \Delta \;\le\; \frac{2^{3/4}\,(1-\eta)^{1/2}\,\sigma}
        {(1+\varepsilon)^{1/4}\sqrt{w(1-w)}}\;n^{-1/4},
    \]
        then $\mathcal R_n(\Delta)\ge\eta$. In particular $\mathcal R_n(\Delta)\ge\tfrac12$ whenever
    \[
        0<\Delta\leq\Delta_0,
        \qquad
        n\;\le\;\frac{2\,\sigma^4}
        {(1+\varepsilon)w^2(1-w)^2\,\Delta^4}.
    \]
\end{theorem}

The proof, together with the two classical information-theoretic facts it relies on
and the explicit two-point construction, is given in
Appendix~\ref{app:minimax-proof}.

The agreement between the criterion-specific lower bound in
Theorem~\ref{thm:sample complexity 2-component} and the minimax lower bound in
Theorem~\ref{thm:minimax-testing} has a direct testing-theoretic explanation.
For the fixed pair $P_r$ and $Q_r$ used in both results, the infimum in Le
Cam's two-point identity is attained by the likelihood-ratio test
\[
    \psi^*=\mathbf 1\left\{
    \sum_{i=1}^n\log\frac{p_r(X_i)}{q_r(X_i)}>0
    \right\}.
\]
Thus, for this pair, the likelihood-based criterion analyzed in
Theorem~\ref{thm:sample complexity 2-component} is precisely the test
underlying the two-point lower bound. Moreover, the mean matching between
$P_r$ and $Q_r$ gives
\[
    D_{\mathrm{KL}}(P_r\|Q_r)=\kappa_w r^4+O(r^6).
\]
Consequently, both the standardized log-likelihood ratio and the
total-variation bound in Le Cam's argument are governed by
$\sqrt{nD_{\mathrm{KL}}(P_r\|Q_r)}$, whose leading scale is
$\sqrt n\,r^2$. Both arguments therefore become nontrivial when $nr^4$ is of
constant order, giving the same critical sample-size exponent $r^{-4}$, or
equivalently $\Delta^{-4}$ for fixed $\sigma$.

The two conclusions nevertheless concern different risks.
Theorem~\ref{thm:sample complexity 2-component} bounds the error of the
likelihood-based criterion under $P_r$, whereas
Theorem~\ref{thm:minimax-testing} bounds the minimum sum of the two testing
errors after restricting the composite model classes to the pair $P_r,Q_r$.
Hence, the agreement of the exponents reflects their common local testing
geometry, but does not by itself provide a matching upper bound or establish
rate optimality of the likelihood-based criterion.

\subsection{Likelihood based lower bound of sample complexity for \texorpdfstring{$k$}{k}--component GMMs}

We extend the analysis in subsection \ref{sec-likeli-sample} to a $k$-component GMM $P$ defined in \eqref{eqn:model}. Similar to the two-component case, our strategy involves constructing a simpler model $Q$ with $(k-1)$ components that matches the first $2k-3$ moments of the true model $P$.

Without loss of generality, we assume $\sigma^2 = 1$ and perform the asymptotic analysis in the regime where the component means are closely located in an interval $[-R, R]$ with $R \ll 1$, i.e., $\mu_i \in [-R, R]$ for all $i = 1,\dots,k$.

Note that a $(k-1)$-component GMM with fixed variance possesses $2(k-1)-1 = 2k-3$ free parameters (mixing weights and means). Consequently, we determine the approximation $Q$ by matching the first $2k-3$ moments of $P$.

\begin{proposition}
\label{prop:k-component}
    Suppose that $\mu_i=Ra_i$ for $i=1,\ldots,k$, where $a_1,\ldots,a_k\in[-1,1]$ are fixed and distinct, and $w_1,\ldots,w_k$ are fixed positive weights satisfying $\sum_{i=1}^k w_i=1$. Define
    \[
        \overline{\nu}=\sum_{i=1}^k w_i\delta_{a_i},
    \]
    and let
    \[
        \overline{\nu}'=\sum_{i=1}^{k-1}\pi_i\delta_{b_i}
    \]
    be the $(k-1)$-point Gaussian quadrature distribution satisfying
    \[
        m_j(\overline{\nu}')=m_j(\overline{\nu}),
        \qquad j=1,\ldots,2k-3.
    \]
    Define the scaled mixing distributions
    \[
        \nu_R=\sum_{i=1}^k w_i\delta_{Ra_i},
        \qquad
        \nu_R'=\sum_{i=1}^{k-1}\pi_i\delta_{Rb_i}.
    \]
    Let $P_R$ and $Q_R$ be the Gaussian mixtures generated by $\nu_R$ and $\nu_R'$, respectively, with common variance $1$ and densities $p_R$ and $q_R$. For $X\sim P_R$, define
    \[
        Y=\log\frac{p_R(X)}{q_R(X)},
    \]
    and let
    \[
        g(x)=\frac{m_{2k-2}(\overline{\nu})-m_{2k-2}(\overline{\nu}')}{(2k-2)!}H_{2k-2}(x),
    \]
    where $H_j$ is the $j$-th probabilists' Hermite polynomial. Then, for $Z\sim\mathcal{N}(0,1)$, we have $\E[g(Z)]=0$ and, as $R\to0$,
    \begin{align*}
        \E[Y] &= \frac{1}{2}\E[g^2(Z)]R^{4k-4}+o(R^{4k-4}),\\
        \operatorname{Var}(Y) &= \E[g^2(Z)]R^{4k-4}+o(R^{4k-4}),\\
        \E\left[|Y-\E[Y]|^3\right] &= \E[|g(Z)|^3]R^{6k-6}+o(R^{6k-6}).
    \end{align*}
\end{proposition}



\begin{theorem}
\label{thm:sample complexity of k-component}
    Let $a_i=-1+\frac{2(i-1)}{k-1}$ for $i=1,\ldots,k$, and consider the GMM
    \[
        P_R=\sum_{i=1}^k w_i\mathcal{N}(Ra_i,1),
    \]
    where $w_1,\ldots,w_k$ are fixed positive weights satisfying $\sum_{i=1}^k w_i=1$, and let $p_R$ be the density of $P_R$. For every $0<\delta<\frac{1}{4}$, there exist constants $N_{k,w,\delta}>0$, $R_0>0$, and $C_{k,w}>0$ such that, for every integer $n\geq N_{k,w,\delta}$, if
    \[
        0<R<\min\left\{R_0,\,C_{k,w}[-\Phi^{-1}(2\delta)]^{\frac{1}{2k-2}}n^{-\frac{1}{4k-4}}\right\},
    \]
    then the $(k-1)$-component GMM $Q_R$ constructed in Proposition \ref{prop:k-component}, with density $q_R$, satisfies
    \[
        \mathbb{P}\left(\frac{1}{n}\sum_{j=1}^n \log p_R(X_j) \leq \frac{1}{n}\sum_{j=1}^n \log q_R(X_j)\right) \geq \delta,
    \]
    where $X_1,\ldots,X_n\stackrel{\mathrm{i.i.d.}}{\sim}P_R$.
\end{theorem}

Since the component means in this construction are equally spaced with $\Delta=2R/(k-1)$, this explicit hard family yields the likelihood-based separation threshold $\Delta=\Omega(n^{-1/(4k-4)})$ and the necessary sample-size scaling $n=\Omega(\Delta^{-(4k-4)})$ for the criterion considered here.

\begin{corollary}
\label{thm:lower bound on sample complexity}
    Under the setting of Theorem \ref{thm:sample complexity of k-component}, let $\Delta=2R/(k-1)$ and $0<\delta<1/4$. For sufficiently small $\Delta$, suppose that $n\geq N_{k,w,\delta}$. If the likelihood-based criterion favors $P_R$ over the $(k-1)$-component GMM $Q_R$ constructed in Proposition \ref{prop:k-component} with probability at least $1-\delta$, then
    \begin{equation}
        \label{eqn:lower bound on sample complexity}
        n\geq c_{k,w}[-\Phi^{-1}(2\delta)]^2\frac{1}{\Delta^{4k-4}},
    \end{equation}
    where $c_{k,w}>0$ depends only on $k$ and the mixing weights. In particular, the sample size required by this likelihood-based criterion for the constructed family satisfies $n=\Omega(\Delta^{-(4k-4)})$ as $\Delta\to0$.
\end{corollary}

This follows immediately from Theorem \ref{thm:sample complexity of k-component} by substituting $R=(k-1)\Delta/2$ and rearranging its resolution condition. Thus, the construction supplies a worst-case example for the likelihood-based comparison used here; it does not by itself constitute an unconditional minimax lower bound over all estimators.

\subsection{A Minimax Lower Bound on the Sample Complexity of Order Estimation for \texorpdfstring{$k$}{k}-Component GMMs}
\label{sec:crl gmm}

We now formalize a minimax lower bound on the sample complexity for estimating the
model order $k$, with $k\geq 2$, of a $k$-component GMM with known common variance. Using Le~Cam's method,
we upgrade the hard-instance bound of Corollary~\ref{thm:lower bound on sample complexity}
to an \emph{unconditional} minimax lower bound that holds over all tests and estimators.
Throughout, for ease of presentation, the common variance is normalized to $\sigma^2=1$.

We denote $\mathcal G_{j,1}$ as the class of GMMs with variance $1$ and $j$ components
\[
\mathcal G_{j,1}:= \{Q : Q \text{ is a GMM with variance $1$}  \text{ and } j \text{ components} \}. 
\]
We note that model order selection for finite mixtures is a nested testing problem, and the nesting is
degenerate: a $(k-1)$-component mixture is a limit of $k$-component mixtures (via a vanishing weight or two merging means), and a $k$-component mixture likewise sits on the
boundary of the $(k+1)$-component class. Over the \emph{unconstrained} classes
$\mathcal G_{k,1}$ and $\mathcal G_{k-1,1}$ the two hypotheses touch, so no separation-free
minimax statement is possible. To address this issue, we hold the minimum separation fixed. For
$\Delta>0$, $R>0$, and weights bounded below by $w_{\min}>0$, let
\[
    \mathcal P_\Delta^{(k)}
    :=\Bigl\{\textstyle\sum_{i=1}^k w_i\,\mathcal N(\mu_i,1):
        \min_{i\neq j}|\mu_i-\mu_j|\ge\Delta,\ \mu_i\in[-R,R],\
        w_i\ge w_{\min}\Bigr\},
\]
and 
\[
 \mathcal Q^{(k-1)}:=\mathcal G_{k-1,1}. 
\]
On these two classes the hypotheses are separated by construction, and the two-point
method applies. More specifically, 
given $X_1,\dots,X_n$ i.i.d.\ from one of the two laws, a test $\psi:\mathbb R^n\to\{0,1\}$
decides between order $k-1$ ($\psi=0$) and order $k$ ($\psi=1$). Define the following \textbf{minimax testing risk}
\begin{equation}
    \label{eqn:minimax-risk-k}
    \mathcal R_n^{(k)}(\Delta)
    :=\inf_{\psi}\Bigl[\
        \sup_{Q\in\mathcal Q^{(k-1)}}\mathbb E_{Q^{\otimes n}}[\psi]
        \;+\;
        \sup_{P\in\mathcal P_\Delta^{(k)}}\mathbb E_{P^{\otimes n}}[1-\psi]\
    \Bigr].
\end{equation}
A value bounded away from $0$ certifies that no procedure can reliably separate model
orders $k$ and $k-1$ at sample size $n$.

\begin{theorem}[Minimax lower bound for model order $k$ vs.\ $k-1$]
\label{thm:minimax-k}
    Fix $k\ge2$, weights $w_1,\dots,w_k>0$, and known variance $1$. Let 
    $a_i=-1+\tfrac{2(i-1)}{k-1}$, set $R=\tfrac{k-1}{2}\Delta$, and let
    $$
    P_R=\sum_{i=1}^k w_i\,\mathcal N(Ra_i,1) \in\mathcal P_\Delta^{(k)}, \,\, Q_R=\sum_{i=1}^{k-1}\pi_i\,\mathcal N(Rb_i,1)\in\mathcal Q^{(k-1)}
    $$ 
    be constructed in Proposition~\ref{prop:k-component}. Then there exists a constant $\kappa_{k,w}>0$, depending only on $k$ and the weights $w_i$, such that, for every $\varepsilon\in(0,1)$, there exists $\Delta_0>0$ for which, whenever $0<\Delta\leq\Delta_0$,
    \begin{equation}
        \label{eqn:minimax-lb-k}
        \mathcal R_n^{(k)}(\Delta)
        \;\ge\;
        1-\sqrt{\tfrac{n}{2}\,D_{\mathrm{KL}}(P_R\|Q_R)}
        \;\ge\;
        1-\sqrt{\tfrac{(1+\varepsilon)\kappa_{k,w}}{2}}\;
        \Delta^{2k-2}\sqrt{n}.
    \end{equation}
    Consequently, for any target level $\eta\in(0,1)$, the minimax risk satisfies
    $\mathcal R_n^{(k)}(\Delta)\ge\eta$ whenever $0<\Delta\leq\Delta_0$ and
    \[
        n\;\le\;\frac{2(1-\eta)^2}{(1+\varepsilon)\kappa_{k,w}}\,
        \frac{1}{\Delta^{4k-4}},
        \qquad\text{equivalently}\qquad
        \Delta\;\le\;\Bigl(\tfrac{2(1-\eta)^2}
        {(1+\varepsilon)\kappa_{k,w}\,n}\Bigr)^{\!\frac{1}{4k-4}}.
    \]
    In particular $\mathcal R_n^{(k)}(\Delta)\ge\tfrac12$ once
    $0<\Delta\leq\Delta_0$ and
    $n\le\bigl(2(1+\varepsilon)\kappa_{k,w}\bigr)^{-1}\Delta^{-(4k-4)}$: no test separates model order $k$
    from $k-1$ at nontrivial power below this sample size.
\end{theorem}

\begin{remark}[Comparison of lower-bound exponents]
\label{rmk:minimax-comparison-k}
    For the constructed family $P_R$ of Theorem~\ref{thm:sample complexity of k-component},
    Corollary~\ref{thm:lower bound on sample complexity} gives the criterion-specific bound
    $\Omega\bigl(\Delta^{-(4k-4)}\bigr)$, while
    Theorem~\ref{thm:minimax-k} shows that the same exponent $4k-4$ governs the
    \emph{unconditional} minimax risk over $\mathcal P_\Delta^{(k)}$ versus
    $\mathcal Q^{(k-1)}$. Thus, the two lower bounds have the same separation
    exponent. This comparison does not provide a success guarantee for the
    likelihood-based criterion above the stated sample-size scale.
\end{remark}

We can also compare this separation exponent with the \textit{Computational Resolution Limit} (CRL) theory established for line spectral estimation (\cite{liu2021mathematical, liu2021theory}). In that framework, the CRL, denoted by $\mathcal{D}_{num}$, represents a minimum separation associated with detecting $k$ sources at a given signal-to-noise ratio (SNR), and scales as
\begin{equation}
    \mathcal{D}_{num} \propto \text{SNR}^{-\frac{1}{2k-2}}.
\end{equation}
In the context of learning GMMs from samples, the ``noise'' in the Fourier measurements is induced by the empirical approximation of the characteristic function. Since the standard deviation of this empirical noise scales as $O(n^{-1/2})$, the effective SNR is heuristically proportional to $\sqrt{n}$. Substituting $\text{SNR}\propto\sqrt{n}$ into the CRL scaling law gives
\begin{equation}
    \Delta_{RL} \propto (\sqrt{n})^{-\frac{1}{2k-2}} = n^{-\frac{1}{4k-4}}.
\end{equation}
This calculation is consistent with the exponent $4k-4$ obtained from the moment-matched hard family. It provides a useful analogy with line spectral estimation, but not an independent minimax characterization of the GMM model-selection problem.

\section{Estimation algorithms based on Fourier data}
\label{sec:fourier}
In this section, we present a method for determining the model order from i.i.d. samples. The proposed algorithm is computationally efficient and its time complexity is linear in the sample size $n$, making it highly scalable. We derive a sufficient recovery guarantee whose dependence on the separation $\Delta$ has the same exponent as the preceding hard-instance result. The guarantee is oracle in the sense that the prescribed threshold depends on the unknown quantities $k$, $w_{\min}$, and $\Delta$. A key novelty of our approach is the utilization of Fourier measurements derived from the samples. Essentially, the algorithm exploits the exponential form of the Fourier measurement (similar to Prony's method introduced in \cite{prony1795essai}) to construct a Hankel matrix that admits a Vandermonde decomposition.
The algorithm proceeds by performing Singular Value Decomposition (SVD) on the Hankel matrix to identify the noise subspace. We distinguish this approach from the algorithm proposed in \cite{qiao2022fourier}, which focuses on learning the means of spherical, equal-weight GMMs in the low-dimensional regime ($d=o(\log k)$). The algorithm therein relies on estimating the Fourier transform of the mixture at carefully chosen frequencies and requires a mean separation of approximately $d/\sqrt{\log k}$ (modulo doubly logarithmic factors).

The Fourier transform for the density in \eqref{eqn:model} is given by
\begin{equation}
\label{eqn:fourier transform}
    \phi(t) = \exp\left(-\frac{\sigma^2 t^2}{2}\right) \sum_{i=1}^k w_i \exp\left(\iota \mu_i t\right),
\end{equation}
which is also known as the characteristic function (CF) in the context of probability theory. An empirical estimate of this is provided by the empirical characteristic function (ECF): 
\[\psi_n(t) = \frac{1}{n}\sum_{j=1}^n \exp\left(\iota X_j t\right).\]
The modulated Fourier measurement of the GMM is then defined as
\begin{equation}
\label{eqn:fourier measurement}
    y_n(t) = \exp\left(\frac{\sigma^2t^2}{2}\right) \frac{1}{n}\sum_{j=1}^n e^{\iota X_j t}.
\end{equation}
This measurement can be regarded as the empirical mixing distribution and it can be shown that:
\begin{proposition}
\label{prop:concentration}
    For any $\epsilon>0$ and fixed $t\in\R$, we have
    \[
        \mathbb{P}\left(|y_n(t) - \mathcal{F}[\nu](t)| \geq \epsilon\right) \leq 4\exp\left\{- \frac{n\epsilon^2}{4\exp(\sigma^2 t^2)}\right\},
    \]
    where $\mathcal{F}[\nu](t) = \sum_{i=1}^k w_i\exp(\iota \mu_i t)$ is the Fourier transform of the mixing distribution $\nu$. 
\end{proposition}
We define the noise term associated with the Fourier measurement as:
    \begin{equation}
    \label{eqn:noise term}
        \epsilon_n(t) = y_n(t) - \mathcal{F}[\nu](t).
    \end{equation}
Estimation of the means $\{\mu_i\}_{i=1}^k$ from $\mathcal{F}[\nu]$ contaminated with noise is known as the Line Spectral Estimation (LSE). The hardness of the LSE can be characterized by the signal-to-noise ratio and the separation distance $\Delta$ of the model. This is quantified by a computational resolution limit introduced in \cite{liu2021mathematical, liu2021theory}.

\subsection{Algorithms for Model Selection and Upper Bound on Sampling Complexity}
We now detail our approach for model selection using Fourier measurements. The Fourier measurement is built on a uniform grid as follows:
    \begin{equation}
    \label{eqn:modulated Fourier measurement}
        y_n(t_q) = \exp\left(\frac{\sigma^2t_q^2}{2}\right) \frac{1}{n}\sum_{j=1}^n e^{\iota X_j t_q} = \sum_{i=1}^k w_i e^{\iota \mu_i t_q} + \epsilon_n(t_q),\ q = 0, \ldots, 2L,
    \end{equation}
where $t_q = -f + q\cdot\frac{f}{L}$. Here, $L \geq k$ represents an a priori upper bound on the true component number $k$, and $f$ denotes the cutoff frequency of the Fourier measurements. For the purpose of model selection, we set
\begin{equation}
\label{eqn:cutoff}
f = \min \left\{\sqrt{\frac{2L-2}{\sigma^2}}, \frac{\pi L}{2R} \right\},
\end{equation}
where the first term balances noise amplification against the frequency bandwidth, while the second term satisfies the Nyquist–Shannon sampling criterion, ensuring the sampling step size does not exceed $\pi / R$.
The Hankel matrix formed by the modulated Fourier measurement is:
    \begin{equation}
        \label{eqn:Hankel matrix}
        H = \begin{bmatrix}
                y_n(t_{0}) & y_n(t_{1}) & \cdots & y_n(t_{L})   \\
                y_n(t_{1}) & y_n(t_{2}) & \cdots & y_n(t_{L+1}) \\
                \vdots     & \vdots     & \ddots & \vdots        \\
                y_n(t_{L}) & y_n(t_{L+1})&\cdots & y_n(t_{2L})
            \end{bmatrix} \in \C^{(L+1)\times (L+1)},
    \end{equation}
and we denote its singular values as $s_1\geq s_2 \geq \cdots \geq s_{L+1}\geq 0$.
In the noiseless case, where $\epsilon_n(t_q) = 0$ for all $q$ in \eqref{eqn:modulated Fourier measurement}, the rank of the Hankel matrix is exactly the model order $k$ for $L \geq k$. Due to the finite sample size $n$ and the resulting noise in the Fourier measurement, the rank-$k$ matrix is perturbed, and the perturbation of the singular values can be estimated quantitatively by the following theorem:
\begin{theorem}
\label{thm:thresholding}
    Suppose that $L \geq k$ and $\mu_i \in [-\frac{(k-1)\pi}{2f}, \frac{(k-1)\pi}{2f}]$ for $i =1,\ldots,k$. Then for any noise threshold $\epsilon < w_{\min}$ and statistical threshold $\delta \in (0, 1)$, if the sample size satisfies
\begin{equation}
            \label{eqn:thresholding sample size}
            n \geq \frac{4\exp(\sigma^2f^2)}{\epsilon^2}\log\left(\frac{4(2L+1)}{\delta}\right),
\end{equation}
we have with probability at least $1-\delta$
        \[
            s_j \leq (L+1)\epsilon, \,\,\, j = k + 1, \cdots L+1. 
        \]
Moreover, if we choose the noise threshold $\epsilon$ such that 
        \begin{equation}
            \label{eqn:thresholding epsilon}
            \epsilon < \frac{w_{\min}\zeta(k)^2(f\Delta)^{2k-2}}
            {2k(\pi L)^{2k-2}(L+1)},
        \end{equation}
    where $\zeta(k)$ is defined in \eqref{eqn:zeta}, then with probability at least $1-\delta$,
        \[
            s_k > (L+1)\epsilon.
        \]
        In this case, the noise singular values and the signal singular values are separated by the threshold $(L+1)\epsilon$. 
\end{theorem}

The above theorem provides a simple thresholding procedure to determine the model order at a prescribed confidence $1-\delta$, which also implies an upper bound on the sample size needed for model selection, as shown in the following corollary.
    \begin{corollary}
    \label{cor:sampling size}
        Consider the GMM $P$ in \eqref{eqn:model}. Suppose that $L\geq k$ and $\mu_i \in \left[-\frac{(k-1)\pi}{2f}, \frac{(k-1)\pi}{2f}\right]$ for $i=1,\ldots,k$. Define
            \[
                \epsilon_*
                =\frac{w_{\min}\zeta(k)^2(f\Delta)^{2k-2}}
                {4k(\pi L)^{2k-2}(L+1)},
            \]
        and suppose that $\epsilon_*<w_{\min}$. For any $\delta\in(0,1)$, if
            \begin{equation}
                n\geq
                \frac{64k^2(\pi L)^{4k-4}(L+1)^2\exp(\sigma^2f^2)}
                {w_{\min}^2\zeta(k)^4(f\Delta)^{4k-4}}
                \log\left(\frac{4(2L+1)}{\delta}\right),
            \end{equation}
        then Algorithm \ref{algo:thresholding}, with threshold $\epsilon=\epsilon_*$, returns the correct model order $k$ with probability at least $1-\delta$.
    \end{corollary}
\begin{proof}
    The choice of $\epsilon_*$ is one half of the upper bound in \eqref{eqn:thresholding epsilon}, and the displayed sample-size condition is exactly \eqref{eqn:thresholding sample size} with $\epsilon=\epsilon_*$. Therefore, Theorem \ref{thm:thresholding} implies that, with probability at least $1-\delta$, the first $k$ singular values exceed $(L+1)\epsilon_*$, whereas the remaining singular values are below this threshold. Hence Algorithm \ref{algo:thresholding} returns $k$.
\end{proof}
The preceding corollary can also be compared directly with the minimax lower
bound. To make this comparison, consider the binary Fourier test
    \[
        \psi_F
        =\mathbf 1\!\left\{s_k(H)>(L+1)\epsilon_*\right\},
    \]
    for testing the class $\mathcal Q^{(k-1)}$ against the $k$-component
    class $\mathcal P_\Delta^{(k)}$. For comparison with
    Theorem~\ref{thm:minimax-k}, normalize the common variance to one and take
    the support bound in $\mathcal P_\Delta^{(k)}$ to satisfy
    $R\leq (k-1)\pi/(2f)$, consistently with Corollary~\ref{cor:sampling size}.

    Under the null hypothesis, the population Hankel matrix has rank at most
    $k-1$. Therefore, on the concentration event used in
    Theorem~\ref{thm:thresholding},
    \[
        s_k(H)\leq(L+1)\epsilon_*,
    \]
    and hence
    \[
        \sup_{Q\in\mathcal Q^{(k-1)}}
        \mathbb P_Q(\psi_F=1)\leq\delta.
    \]
    Under the alternative hypothesis, Corollary~\ref{cor:sampling size} gives
    \[
        \sup_{P\in\mathcal P_\Delta^{(k)}}
        \mathbb P_P(\psi_F=0)\leq\delta
    \]
    whenever
    \[
        n\geq
        C_{k,L,f}\,w_{\min}^{-2}\Delta^{-(4k-4)}
        \log\!\left(\frac{C_L}{\delta}\right),
    \]
    where $C_{k,L,f}>0$ and $C_L>0$ are independent of $n$ and $\Delta$.
    Consequently, the worst-case sum of the Type~I and Type~II errors of
    $\psi_F$ is at most $2\delta$. On the other hand,
    Theorem~\ref{thm:minimax-k} shows that the minimax testing risk remains
    bounded away from zero when
    \[
        n\leq c_{k,w}\Delta^{-(4k-4)}
    \]
    for a sufficiently small constant $c_{k,w}>0$.

    Thus, for fixed $k,L,f,w_{\min}$ and a fixed confidence level, the lower
    and upper bounds have the same dependence on $\Delta$. The oracle Fourier
    test is therefore minimax-rate optimal with respect to the separation
    exponent $4k-4$. This statement concerns only the exponent in $\Delta$; it
    does not claim optimal constants, optimal dependence on the remaining
    parameters, or adaptation to unknown $k$, $w_{\min}$, and $\Delta$.
Based on the corollary, the thresholding algorithm can be summarized as follows.

    \begin{algorithm}[H]
        \label{algo:thresholding}
        \caption{Model Selection by Thresholding}
        \Input{samples $X_1, \cdots, X_n$, variance $\sigma^2$, cutoff frequency $f$, Hankel matrix size $L+1$, thresholding term $\epsilon$}

        $t_q \gets -f + q \frac{f}{L}$ for $q = 0,1,\cdots,2L$\;
        
        $y_n(t_q) \gets \exp\left(\sigma^2 t_q^2 / 2\right)\sum_{j=1}^n \exp(\iota X_j t_q)/n$ for $q = 0, 1, \cdots, 2L$\;

        $H \gets \eqref{eqn:Hankel matrix}$\;

        $s_1 \geq s_2 \geq \cdots \geq s_{L+1} \gets$ singular values of $H$\;

        \Output{estimated model order $\widehat{k} \gets \max\{i: s_i > (L+1)\epsilon\}$}

    \end{algorithm}

In the special case of distinguishing a one-component Gaussian from a two-component GMM with separation distance $\Delta$, the algorithm can be simplified by eliminating the hyperparameter $\epsilon$. We consider the following simplified rule:

    \begin{algorithm}[H]
    \label{algo: OneTwoModel}
        \caption{Model Selection between One-component Gaussian or Two-component GMMs}
        \Input{samples $X_1, \cdots, X_n$, variance $\sigma^2$, cutoff frequency $f$, separation distance $\Delta$}
        $\widetilde{X}_j \gets X_j - \frac{1}{n}\sum_{i=1}^n X_i$ for $j=1,\ldots,n$\;
        
        $y_n \gets \exp\left(\sigma^2f^2/2\right)\sum_{j=1}^n \exp\left(\iota \widetilde{X}_j f \right)/n$\;

        \uIf{$y_n + \bar{y}_n > 2\cos^2(\Delta f / {4})$}{
        $\widehat{k} \gets 1$ \;
        }
        \uElse{
        $\widehat{k} \gets 2$ \;
            }
        \Output{estimated model order $\widehat{k}$.}
    \end{algorithm}
    

One drawback of Algorithm \ref{algo:thresholding} is that it requires a hyperparameter $\epsilon < w_{\min}$ and the sample size $n$ satisfying $n = O\left(\frac{1}{\epsilon^2 \Delta^{4k-4}}\right)$. To avoid the selection of $\epsilon$, we propose an alternative method based on the Hankel structure inherent in the Fourier measurement. It is based on the fact that in the noiseless case, the Hankel matrix has rank $k$, which is equivalent to 
\[
    \frac{s_k}{s_{k+1}} = + \infty.
\]
Based on this idea, the alternative method uses a singular value ratio measure and is summarized below:

\begin{algorithm}
\label{algo:singular value ratio}
    \caption{Singular Value Ratio Model Selection}
    \Input{samples $X_1, \cdots, X_n$, variance $\sigma^2$, cutoff frequency $f$, Hankel matrix size $L+1$}

    $t_q \gets -f + q \frac{f}{L}$ for $q = 0,1,\cdots,2L$\;
        
    $y_n(t_q) \gets \exp\left(\sigma^2 t_q^2 / 2\right)\sum_{j=1}^n \exp(\iota X_j t_q)/n$ for $q = 0, 1, \cdots, 2L$\;

    $H \gets \eqref{eqn:Hankel matrix}$\;

    $s_1 \geq s_2 \geq \cdots \geq s_{L+1} \gets$ singular values of $H$\;

    \Output{estimated model order $\widehat{k} \gets \arg\max_{1\leq i\leq L} s_i/s_{i+1}$.}
\end{algorithm} 

\subsection{Improved Sample Complexity in the Non-Worst Case}
\label{sec:local-rates}

The global exponent $4k-4$ is attained by the hard configuration in which all
$k$ atoms are closely spaced within a window of width $(k-1)\Delta$, so that all
$k-1$ relevant Vandermonde factors degenerate simultaneously. When only a subset
of the atoms approaches collision, the Fourier analysis yields a sharper
sufficient sample-size bound. In this subsection we show that this bound depends
on the number of atoms that collide rather than on the total number of atoms $k$.
When only a single pair approaches collision, the sufficient dependence improves
from $O(\Delta^{-(4k-4)})$ to $O(\Delta^{-4})$, with a $\Delta$-exponent
independent of $k$.

The regime we analyze is the
following.

\begin{definition}[Single-collision configuration]
\label{def:single-collision}
    Fix a resolution constant $0<c_0\leq(k-1)\pi$ that is independent of $\Delta$, $n$, and $L$.
    After relabelling, exactly one pair of nodes is merged,
    \[
        |\mu_1-\mu_2| \;=\; \Delta, \qquad 0<f\Delta<c_0,
    \]
    while every remaining separation is resolved,
    \[
        c_0 \;\le\;
        \min_{\substack{i\neq j\\ \{i,j\}\neq\{1,2\}}}
        f\,|\mu_i-\mu_j|,
        \qquad\text{and}\qquad
        \max_{i\neq j} f\,|\mu_i-\mu_j| \;\le\; (k-1)\pi,
    \]
    where the lower bound on the remaining separations is understood as vacuous
    when $k=2$. The upper bound is forced by $\mu_i\in\bigl[-\tfrac{(k-1)\pi}{2f},
    \tfrac{(k-1)\pi}{2f}\bigr]$.
\end{definition}

Under this configuration exactly two of the $k$ component means collide into a single location, while the remaining $k-2$ means stay separated by order-one gaps. 

\begin{theorem}[Thresholding under a single collision]
\label{thm:thresholding-single}
    Suppose that $L \geq k$ and $\mu_i \in \left[-\frac{(k-1)\pi}{2f},
    \frac{(k-1)\pi}{2f}\right]$ for $i=1,\ldots,k$, in the single-collision
    configuration of Definition~\ref{def:single-collision}. Then for any noise
    threshold $\epsilon<w_{\min}$ and statistical threshold $\delta\in(0,1)$, if the
    sample size satisfies
    \begin{equation}
        \label{eqn:thresholding-single sample size}
        n \geq \frac{4\exp(\sigma^2f^2)}{\epsilon^2}
        \log\left(\frac{4(2L+1)}{\delta}\right),
    \end{equation}
    we have with probability at least $1-\delta$
    \[
        s_j \leq (L+1)\epsilon,\qquad j=k+1,\cdots,L+1.
    \]
    Moreover, if the noise threshold $\epsilon$ is chosen such that
    \begin{equation}
        \label{eqn:thresholding-single epsilon}
        \epsilon \ <\ \frac{w_{\min}\,c_0^{2(k-2)}\,(f\Delta)^{2}}
        {2k\,(\pi L)^{2k-2}\,(L+1)},
    \end{equation}
    then with probability at least $1-\delta$,
    \[
        s_k > (L+1)\epsilon.
    \]
    In this case the noise singular values and the signal singular values are
    separated by the threshold $(L+1)\epsilon$.
\end{theorem}

\begin{corollary}[Sample size under a single collision]
\label{cor:sampling-single}
    Consider the GMM $P$ in~\eqref{eqn:model}. Suppose that $L\geq k$ and
    $\mu_i \in \left[-\frac{(k-1)\pi}{2f}, \frac{(k-1)\pi}{2f}\right]$ for
    $i=1,\ldots,k$, in the single-collision configuration of
    Definition~\ref{def:single-collision}. Define
    \[
        \epsilon_*
        =\frac{w_{\min}\,c_0^{2(k-2)}\,(f\Delta)^{2}}
        {4k\,(\pi L)^{2k-2}\,(L+1)},
    \]
    and suppose that $\epsilon_*<w_{\min}$. For any $\delta\in(0,1)$, if
    \begin{equation}
        \label{eqn:sampling-single}
        n\ \geq\
        \frac{64\,k^2\,(\pi L)^{4k-4}\,(L+1)^2\exp(\sigma^2f^2)}
        {w_{\min}^2\,c_0^{4(k-2)}\,(f\Delta)^{4}}
        \log\left(\frac{4(2L+1)}{\delta}\right),
    \end{equation}
    then Algorithm~\ref{algo:thresholding}, with threshold $\epsilon=\epsilon_*$,
    returns the correct model order $k$ with probability at least $1-\delta$.
\end{corollary}

A matching local lower-bound exponent can also be obtained by a direct Le~Cam
two-point argument, without introducing a separate theorem. To see this, keep the
remaining $k-2$ resolved components fixed and write $\alpha=w_1+w_2$. Let
$P_{\mathrm{close}}$ be the conditional two-component mixture formed by the close
pair, and let $Q_{\mathrm{close}}$ be the single Gaussian with variance $\sigma^2$
and mean $(w_1\mu_1+w_2\mu_2)/\alpha$. Replacing the close pair by
$\alpha Q_{\mathrm{close}}$ produces a $(k-1)$-component mixture $Q_\Delta$;
denote the original $k$-component mixture by $P_\Delta$.
By joint convexity of the Kullback--Leibler divergence,
\[
    D_{\mathrm{KL}}(P_\Delta\|Q_\Delta)
    \leq \alpha
    D_{\mathrm{KL}}(P_{\mathrm{close}}\|Q_{\mathrm{close}})
    =O(\Delta^4).
\]
Le~Cam's method therefore keeps the testing risk bounded away from zero whenever
$n\leq c\Delta^{-4}$ for a sufficiently small constant $c>0$. Thus, over the corresponding single-collision class, the
lower-bound exponent agrees with the sufficient exponent in
Corollary~\ref{cor:sampling-single}.

The proofs of the two results, together with the underlying singular-value estimate
of Lemma~\ref{lem:sigmak-single}, are deferred to Appendix~\ref{app:local-rates}.

\paragraph{The general collision case}
The single-collision result is the first nontrivial case of a general principle:
the sufficient exponent is controlled by the size of the largest merging cluster. Suppose that $m$
of the $k$ means merge into a common cluster of width $\Delta$ while the remaining
$k-m$ means stay resolved. Then the argument in the proof of Lemma~\ref{lem:sigmak-single} gives
\begin{equation}
    \label{eqn:general-cluster}
    s_k(Y)\ \ge\
    \frac{w_{\min}\,c_0^{2(k-m)}\,(f\Delta)^{2(m-1)}}
    {k\,(\pi L)^{2k-2}}.
\end{equation}
This interpolates between the two extremes: $m=2$ (a single collision) recovers
Lemma~\ref{lem:sigmak-single} with $f\Delta$-exponent $2$, while $m=k$ (all means
merged) recovers the worst-case bound with $f\Delta$-exponent $2k-2$ and no resolved
factor. Propagating \eqref{eqn:general-cluster} through
Theorem~\ref{thm:thresholding-single} and Corollary~\ref{cor:sampling-single} yields
the sufficient sample-size dependence
\[
    n=O\!\left(\Delta^{-4(m-1)}\right),
\]
whose exponent depends only on the size $m$ of the merging cluster, not on the
total order $k$.

\begin{remark}[Improved sufficient bounds away from the fully clustered case]
\label{rmk:local-rate}
    In the fully clustered configuration ($m=k$), all $k$ means lie within a
    window of width $(k-1)\Delta$, producing the exponent
    $(f\Delta)^{2k-2}$ and the sufficient sample-size dependence
    $O(\Delta^{-(4k-4)})$ in Theorem~\ref{thm:thresholding} and
    Corollary~\ref{cor:sampling size}. In the single-collision configuration
    ($m=2$), the exponent drops to $(f\Delta)^2$, and
    Corollary~\ref{cor:sampling-single} gives, for fixed remaining parameters,
    \[
        n=O(\Delta^{-4}),
    \]
    whose $\Delta$-exponent is independent of $k$. More generally,
    \eqref{eqn:general-cluster} gives the sufficient dependence
    $O(\Delta^{-4(m-1)})$ for a merging cluster of size $m$.
\end{remark}

\subsection{Mixing Distribution Estimation by Fourier Measurement}
The Fourier measurement can also be used to estimate the mixing distribution of the GMMs and is closely related to line spectral estimation. In our algorithm, we use the widely used MUltiple SIgnal Classification (MUSIC) algorithm to estimate the component means. The MUSIC algorithm is a widely utilized technique in frequency estimation, spectral analysis, and radar signal processing, renowned for its high-resolution parameter estimation capabilities. Essentially, the MUSIC algorithm exploits the exponential form of the signals (similar to Prony's method \cite{prony1795essai}) to construct a Hankel matrix that admits a Vandermonde decomposition.
The algorithm proceeds by performing SVD on the Hankel matrix to identify the noise subspace. Subsequently, it formulates an imaging function (denoted as $\mathcal{J}(\mu)$ in the algorithm) by computing a noise-space corn n function. In the noiseless scenario, the imaging function exhibits peaks precisely at the set of Gaussian means $\{\mu_j\}_{1\leq j \leq k}$. In the presence of noise, the algorithm determines the number of Gaussian means by identifying the number of local maxima in the imaging function and estimates the set of means based on the locations of these maxima. The weight of each Gaussian component can be then resolved by the quadratic programming by finding weights satisfying:

    \begin{align}\label{eqn:weight solver}
        &\text{minimize } \sum_{q=0}^{2L}\norm{\sum_{i=1}^{k} w_i \exp(\iota \mu_i t_q) -  y_n(t_q)}_2^2, \nonumber \\
        &\text{subject to } w_i \geq 0, \quad \sum_{i=1}^{\hat{k}} w_i = 1. 
    \end{align}
The details of the MUSIC algorithm can be found in Appendix \ref{app:music}.

\begin{algorithm}
\label{algo:music-based mixing distribution estimation}
    \caption{MUSIC-based Mixing Distribution Estimation}
    \Input{samples $X_1, \cdots, X_n$, variance $\sigma^2$, model order $k$, cutoff frequency $f$, Hankel matrix size $L+1$}
    
    $t_q \gets -f + q \frac{f}{L}$ for $q = 0,1,\cdots,2L$\;
        
    $y_n(t_q) \gets \exp\left(\sigma^2 t_q^2 / 2\right)\sum_{j=1}^n \exp(\iota X_j t_q)/n$ for $q = 0, 1, \cdots, 2L$\;

    $H \gets \eqref{eqn:Hankel matrix}$\;

    $\mathcal{J}(\mu) \gets $ MUSIC algorithm with Hankel matrix $H$ and source number $k$\;

    $\{\hat{\mu}_i\}_{i=1}^k \gets k$ largest local maxima of $\mathcal{J}(\mu)$\;

    $\{\hat{w}_i\}_{i=1}^k \gets $ quadratic programming \eqref{eqn:weight solver}\;

    \Output{mixing distribution $\sum_{i=1}^k \hat{w}_i\delta_{\hat{\mu}_i}$.}
\end{algorithm}
The numerical results of the Algorithm \ref{algo:music-based mixing distribution estimation} and its comparison with the EM algorithm are shown in Section \ref{subsec: numerical mixing distribution estimation}.

\section{Numerical Experiments}
\label{sec:numerical}
\subsection{Model Selection and Empirical Transition}
In this section, we investigate the performance of Algorithm \ref{algo:singular value ratio} and compare it with the commonly used model selection methods based on the information criteria. Specifically, we evaluate and compare the estimation results on two-component, three-component, and four-component equally weighted GMMs with small separation distances. In the experiment, all Gaussian components share the common variance $1.0$. 

\begin{figure}[hbt!]
\label{fig:phase transition}
    \centering
    \subfigure[Result of $2$-component GMMs]{\includegraphics[width=0.8\textwidth]{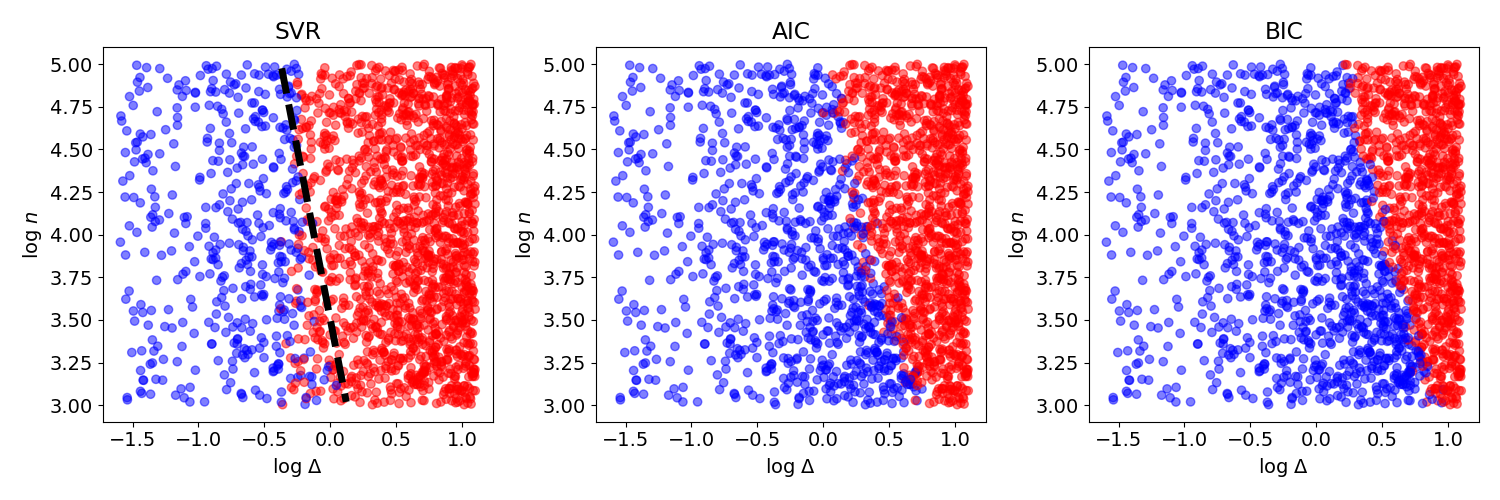}}

    \vskip\baselineskip 
    \subfigure[Result of $3$-component GMMs]{\includegraphics[width=0.8\textwidth]{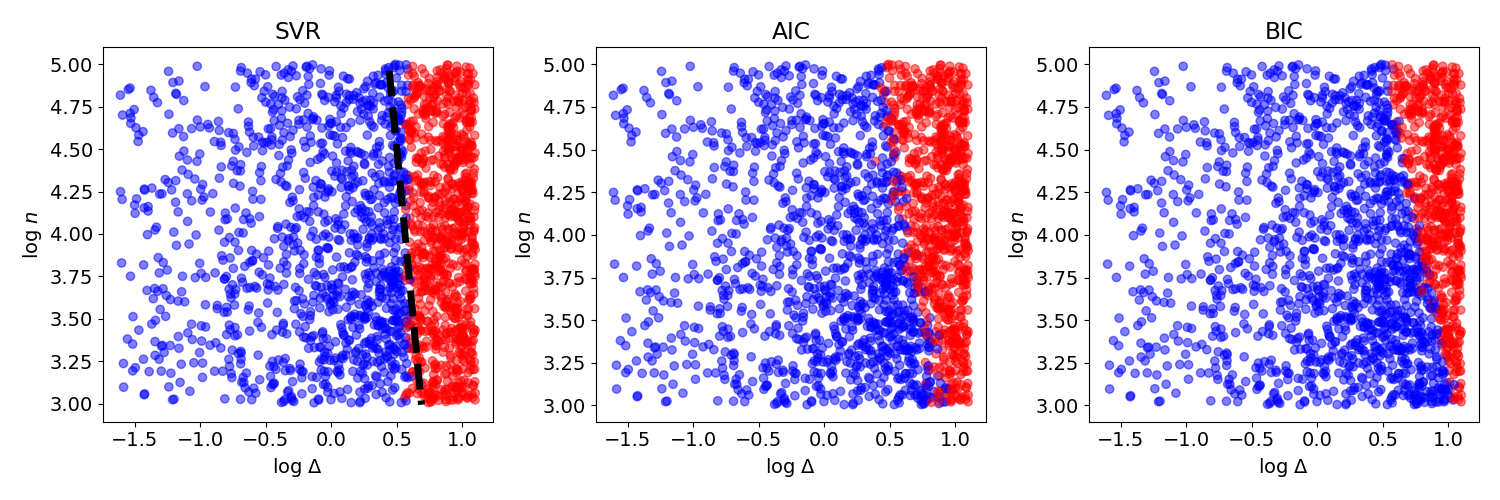}}

    \vskip\baselineskip 
    \subfigure[Result of $4$-component GMMs]{\includegraphics[width=0.8\textwidth]{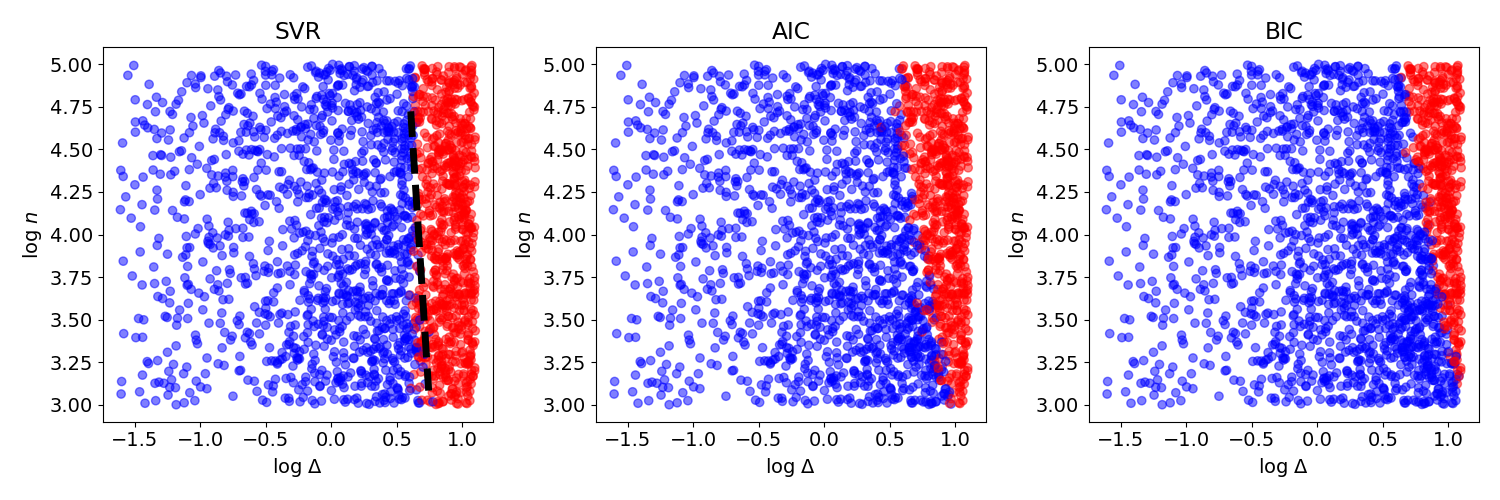}}

    \caption{Empirical transition of different model selection methods with respect to the logarithmic separation distance $\log_{10}\Delta$ and $\log_{10}(n)$. Left: Algorithm \ref{algo:singular value ratio}. Middle: AIC. Right: BIC. The red dots denote correct estimates of the model order, while the blue dots denote incorrect estimates. The slope of the black dashed reference line is $4-4k$ in each SVR plot.}
    \label{fig:three_figures}
\end{figure}

The experiments are designed as follows. For $2$-, $3$- and $4$-component GMMs, we uniformly sample $2,000\ (\log_{10}(n), \Delta)$ pairs from the domain $[3.0, 5.0] \times [0.2, 3.0]$.  We draw $n$ i.i.d. samples from the mixture model with separation distance $\Delta$ and apply Algorithm \ref{algo:singular value ratio} along with AIC and BIC for model selection. For the $k$-component GMM, the inputs of Algorithm \ref{algo:singular value ratio} are $f = \sqrt{2k-2}, L = k+1$, which allows the estimated model order ranging from $1$ to $k+1$. For AIC and BIC, the model is estimated by the EM algorithm with model order ranging from $1$ to $k+1$. The EM algorithm terminates after $5,000$ iterations or the log likelihood increases less than $1\times 10^{-5}$. The results are shown in Figure \ref{fig:phase transition}

The results reveal an empirical transition for all three methods. The proposed method demonstrates a more favorable success rate compared to the information criteria, especially in the small-sample scenarios. It is worth mentioning that Algorithm \ref{algo:singular value ratio} proceeds much faster than the information criteria since both AIC and BIC require separate mixture fitting for each candidate model order, whereas Algorithm \ref{algo:singular value ratio} only necessitates computing the singular value decomposition of the constructed Hankel matrix. The detailed comparison on efficiency can be found in the next subsection. Motivated by the exponent appearing in both the hard-instance and oracle upper-bound analyses, we use the reference scaling
\[
    n \propto \frac{1}{\Delta^{4k-4}}.
\]
Taking logarithms gives a reference line with slope $-(4k-4)$ in the $\log \Delta$ versus $\log n$ plane. The empirical boundary in Figure \ref{fig:three_figures} is broadly consistent with this slope. This numerical agreement illustrates the predicted separation dependence in the tested settings, but does not prove optimality.
\subsection{Mixing Distribution Estimation}
\label{subsec: numerical mixing distribution estimation}
In this part, we conduct numerical experiments to compare the efficiency and accuracy of Algorithm \ref{algo:music-based mixing distribution estimation} with the EM algorithm. To compare the performance of the two algorithms, we plot out the 1-Wasserstein distance error of mixing distribution defined as 
\[
    W_1(\nu, \tilde\nu) = \int \abs{F_\nu(t) - F_{\hat\nu}(t)} d t,
\]
where $\hat \nu = \sum_{i=1}^{\hat k} \hat \pi_i \delta_{\hat \mu_i}$ is the estimated mixing distribution and $F_\nu, F_{\hat\nu}$ denote the cumulative distribution function (CDF) of discrete measures $\nu, \hat\nu$. The average running time of each trial is included to compare the computational cost.


In the first experiment, we consider the following three Gaussian mixtures:
\begin{enumerate}
    \item Two-component GMM: $\frac{1}{2}\mathcal{N}(-0.5, 1.0) + \frac{1}{2}\mathcal{N}(-0.5, 1.0)$;
    \item Three-component GMM: $\frac{1}{3}(\mathcal{N}(-0.5, 1.0) + \mathcal{N}(0.0, 1.0) + \mathcal{N}(-0.5,1.0))$;
    \item Four-component GMM:
    \[
        \frac{1}{4}\bigl(\mathcal{N}(-0.75, 1.0)+\mathcal{N}(-0.25,1.0)
        +\mathcal{N}(0.25, 1.0)+\mathcal{N}(0.75,1.0)\bigr).
    \]
\end{enumerate}
For the EM algorithm, we randomly select two samples as the initial guess of means and set the initial variance equal to the standard deviation of the samples. The algorithm terminates when the log-likelihood increase is less than $1\times10^{-5}$ or after $5,000$ iterations. For Algorithm \ref{algo:music-based mixing distribution estimation}, the inputs are set as $k$, $f = \sqrt{2k-2}$, and $L=k+1$, where $k$ is the number of components.
\begin{figure}[hbt]
    \centering
    \includegraphics[width=\linewidth]{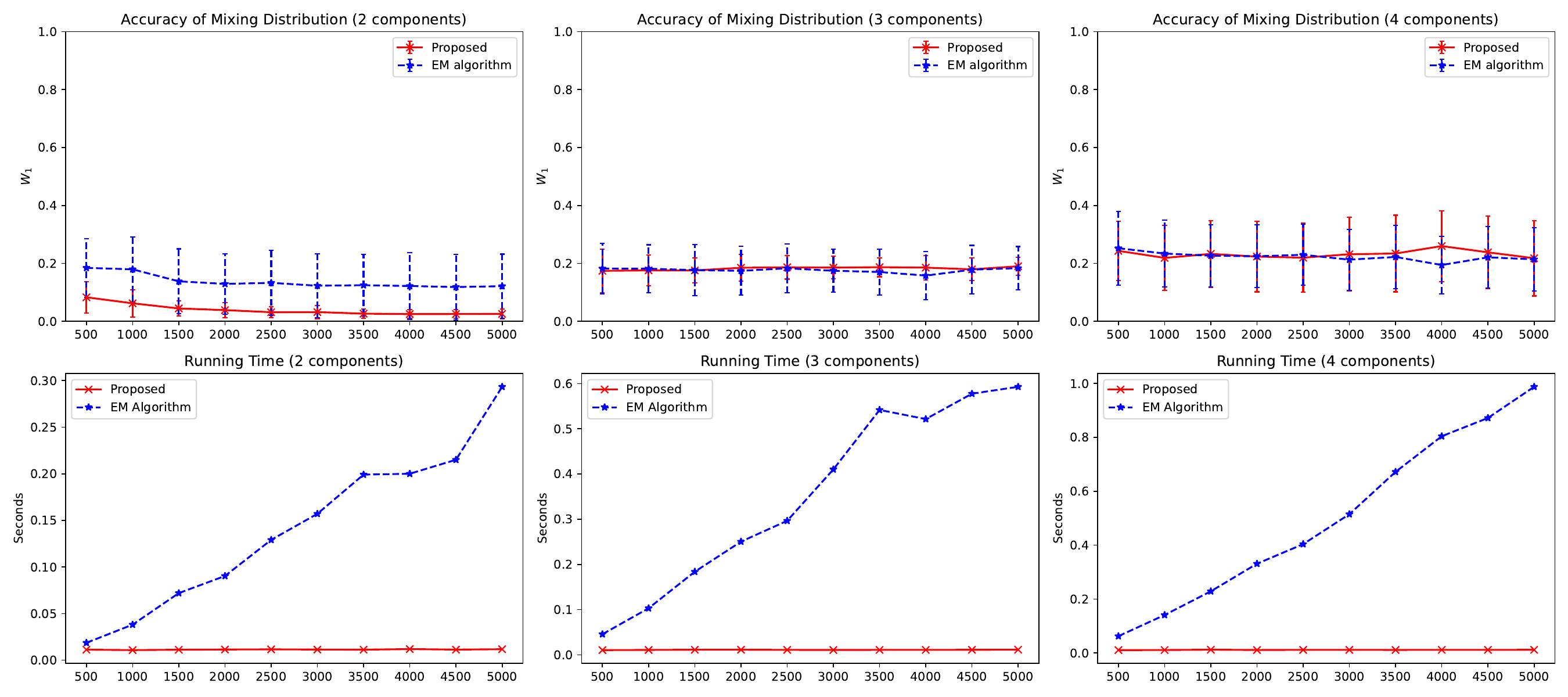}
    \caption{Comparison with the EM algorithm for 2-, 3-, and 4-component GMMs. For each sample size, we conducted 100 trials. The upper plots show the accuracy of mixing distribution estimation; the lower plots show the average running time of each trial.}
    \label{fig:vsEM known}
\end{figure}

For each sample size, 100 trials are conducted and the results are shown in Figure \ref{fig:vsEM known}. When the separation distance is large (see the 2-component mixture), our algorithm performs better than the EM algorithm in accuracy and efficiency across all sample sizes. When the components are closer (see the 3-component and 4-component mixtures), the accuracies are comparable, but our algorithm is much faster than EM, especially for large sample sizes. This is because the EM algorithm accesses all the samples at each iteration, whereas our algorithm uses them only to compute the Fourier data $y_n(t_q)$ in the first step of Algorithm \ref{algo:music-based mixing distribution estimation}.

\begin{figure}[hbt]
    \centering    \includegraphics[width=\linewidth]{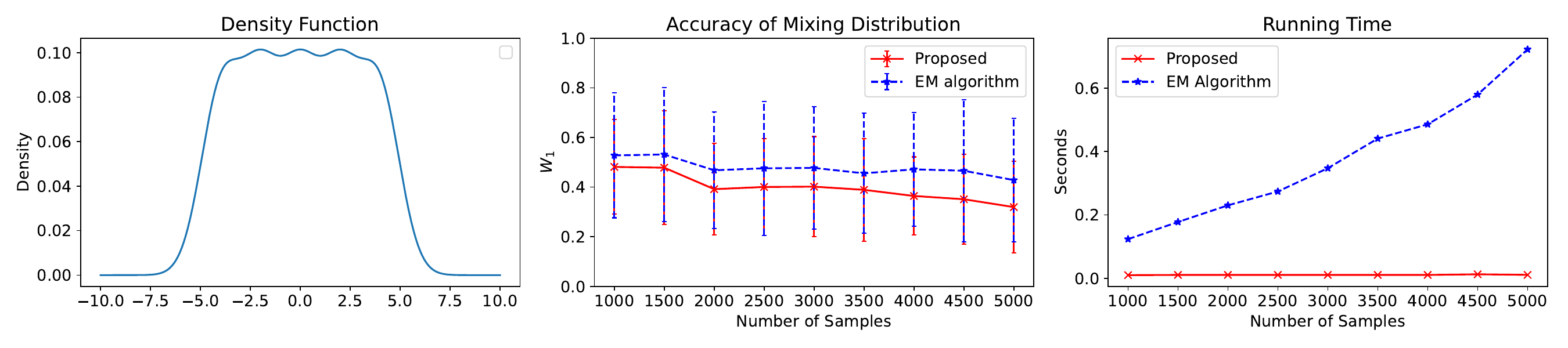}
    \caption{Comparison with the EM algorithm (5 components). For each sample size, we conduct 100 trials. The left plot shows the probability density function of the equally weighted distribution. The middle plot shows the accuracy of mixing distribution estimation; the right plot shows the average running time of each trial.}
    \label{fig: multiple}
\end{figure}

Next, we consider a more challenging scenario involving a Gaussian mixture with 5 components with different weights. The mixing distribution in each trial satisfies
\[
    \nu = w_1\delta_{-4} + w_2\delta_{-2} + w_3\delta_0 + w_4\delta_2 + w_5\delta_4
\]
and the weights $(w_1, w_2, w_3, w_4, w_5)$ are drawn from a Dirichlet distribution with parameter $(1,1,1,1,1)$. The shared variance of Gaussian components is set as $1.0$.
For Algorithm \ref{algo:music-based mixing distribution estimation}, we set $k=5$, $f = \sqrt{2}$, and $L=6$. For the EM algorithm, we terminate when either the log-likelihood increment is less than $1\times 10^{-6}$ or the number of iterations reaches $5,000$. Figure~\ref{fig: multiple} shows the density function of the equally weighted model alongside the numerical results. Our algorithm demonstrates superior performance in terms of both accuracy and running time, especially for larger sample sizes.

It is important to note that Algorithm \ref{algo:music-based mixing distribution estimation} can also be applied for model selection in this example. If the number of local maxima in the MUSIC imaging function is fewer than 5, it suggests that the available sample size $n$ is below the critical sample complexity $N^*$ required to resolve the specific separation distance of those components. Consequently, with a limited number of samples, one can only reliably recover a lower model order.

\clearpage
\section{Limitations and Future Work}
We acknowledge several limitations of the present analysis. First, the likelihood-based lower bounds are obtained from explicit families of closely spaced mixtures and remain specific to the criterion considered here. The subsequent Le~Cam arguments provide procedure-independent minimax lower bounds over the parameter classes stated in Section~\ref{sec:crl}. However, these results do not provide a success guarantee for the likelihood-based criterion above the corresponding sample-size scale, and the scopes of the minimax lower bounds and the Fourier upper bounds must be kept distinct.

Second, the sufficient guarantee for the Fourier thresholding method is oracle in nature: its prescribed threshold depends on the unknown model order $k$, minimum weight $w_{\min}$, and separation $\Delta$. The singular-value-ratio procedure used in the numerical experiments avoids this explicit threshold, but the current theory does not provide it with a corresponding finite-sample recovery guarantee. Developing a fully data-driven threshold or a rigorous analysis of the singular-value-ratio rule is an important direction for future work.

Third, the present framework is restricted to one-dimensional GMMs with a known common variance $\sigma^2$. Unknown or component-dependent variances introduce additional nuisance parameters and may change both the hard-instance construction and the Fourier deconvolution step. Extending the analysis to high-dimensional mixtures is also challenging and may require dimensionality reduction techniques such as PCA (see \cite{vempala2002spectral}) or random projection (see \cite{sanjeev2001learning}), together with adaptations of the Fourier-based approach.

Finally, the numerical experiments illustrate the predicted separation dependence and favorable computational performance only for the tested configurations. They do not establish optimality or robustness under model misspecification, very small mixing weights, or unknown variances. Broader empirical evaluation and theoretical robustness guarantees remain topics for future study.

\clearpage
\appendix
\section{Proofs and Auxiliary Results}
\label{sec:appendix}
\subsection{MUSIC Algorithm}
\label{app:music}
Consider the following signal sampled on the equally-spaced grid:
    \[y(t_q) = \sum_{i=1}^k w_i \exp(\iota \mu_i t_q) + \epsilon_n(t_q), \quad q = 0, \ldots, 2L, \]
    where $t_q = -f + q\frac{f}{L}$. We construct the Hankel matrix $H$ as in \eqref{eqn:Hankel matrix} and then define
    \[
        \phi(\mu) = \begin{bmatrix}
            1 & \exp(\iota \mu \frac{f}{L}) & \exp(\iota 2 \mu \frac{f}{L}) & \cdots & \exp(\iota L\mu\frac{f}{L})
        \end{bmatrix}^\mathrm{T}.
    \]
The MUSIC algorithm can be summarized as follows
\begin{algorithm}
    \label{algo:MUSIC}
    \caption{MUltiple SIgnal Classification (MUSIC) Algorithm}
    \Input{Hankel matrix $H$, source number $k$}
     Perform the SVD of $H$:
     \[
        H = \begin{bmatrix}
            U_1 & U_2
        \end{bmatrix}\diag{s_1, \cdots, s_k, \cdots}
        \begin{bmatrix}
            V_1 & V_2
        \end{bmatrix}^*,
     \]
     where $U_1 \in \mathbb{C}^{(L+1) \times k}$\;

     $\mathcal{J}(\mu) \gets {\norm{\phi(\mu)}_2} / {\norm{U_2^* \phi(\mu)}_2}$ for $\mu \in [-\frac{\pi L}{f}, \frac{\pi L}{f}]$\;

     \Output{the largest $k$ local maxima of $\mathcal{J}(\mu)$.}
\end{algorithm}

\subsection{Proof of Proposition \ref{prop:2-component analysis}}

\begin{proof}
For brevity, write $m=m_r$, $p=p_r$, and $q=q_r$, and set
$z=(x-m)/\sigma$. Since $m=(1-2w)r\sigma$, the likelihood ratio is
\begin{align*}
    S_r(z):=\frac{p(x)}{q(x)}
    &=w\exp\bigl(-2(1-w)rz-2(1-w)^2r^2\bigr)\\
    &\quad +(1-w)\exp\bigl(2wrz-2w^2r^2\bigr).
\end{align*}
Using the generating function
\[
    \exp(tz-t^2/2)=\sum_{j=0}^{\infty}\frac{H_j(z)}{j!}t^j,
\]
we obtain
\begin{equation}
\label{eqn:two-component-ratio-expansion}
    S_r(z)=1+g(z)r^2+h(z)r^3+O(r^4),
\end{equation}
where
\[
    g(z)=2w(1-w)H_2(z),
    \qquad
    h(z)=\frac{4}{3}w(1-w)(2w-1)H_3(z).
\]
Here and below, the remainders are understood after integration against the
standard Gaussian density. More precisely, for every fixed finite $s\geq1$,
the remainder in \eqref{eqn:two-component-ratio-expansion} is $O(r^4)$ in
$L^s(\mathcal N(0,1))$. This follows directly from the generating function and
the Gaussian exponential moments. The logarithmic expansions below follow by
splitting the integral into $|z|\leq r^{-1/2}$, where $S_r(z)-1=o(1)$ uniformly,
and its complement, whose contribution is smaller than any power of $r$ by the
Gaussian tail bound.

Let $u_r=S_r-1$. Since $p$ and $q$ are probability densities,
\[
    \E[u_r(Z)]=\int_{\mathbb R}(p(x)-q(x))\,dx=0,
    \qquad Z\sim\mathcal N(0,1).
\]
Moreover, Hermite orthogonality gives
\[
    \E[g(Z)h(Z)]=0.
\]
Consequently,
\begin{equation}
\label{eqn:two-component-u-square}
    \E[u_r^2(Z)]=\E[g^2(Z)]r^4+O(r^6),
    \qquad
    \E[|u_r(Z)|^3]=O(r^6).
\end{equation}

By the change-of-measure formula and
$(1+u)\log(1+u)=u+u^2/2+O(|u|^3)$,
\begin{align*}
    \E[Y]
    &=\E\bigl[S_r(Z)\log S_r(Z)\bigr]\\
    &=\frac12\E[u_r^2(Z)]+O\bigl(\E[|u_r(Z)|^3]\bigr)\\
    &=\frac12\E[g^2(Z)]r^4+O(r^6).
\end{align*}
Similarly,
\begin{align*}
    \E[Y^2]
    &=\E\bigl[S_r(Z)(\log S_r(Z))^2\bigr]\\
    &=\E[u_r^2(Z)]+O\bigl(\E[|u_r(Z)|^3]\bigr)
      =\E[g^2(Z)]r^4+O(r^6).
\end{align*}
Since $(\E[Y])^2=O(r^8)$, it follows that
\[
    \operatorname{Var}(Y)=\E[g^2(Z)]r^4+O(r^6).
\]

The same truncation argument and \eqref{eqn:two-component-ratio-expansion}
show that, for $X\sim P_r$ and $Z\sim\mathcal N(0,1)$,
\[
    r^{-2}\log S_r\left(\frac{X-m}{\sigma}\right)
    \longrightarrow g(Z)
\]
in distribution and in the third absolute moment. Since $\E[Y]=O(r^4)$, we obtain
\[
    \E[|Y-\E[Y]|^3]
    =\E[|g(Z)|^3]r^6+o(r^6).
\]
Finally, $g(Z)=2w(1-w)(Z^2-1)$ and H\"older's inequality yield
\[
    \E[|Z^2-1|^3]\leq\E[(Z^2-1)^4]^{3/4}=60^{3/4}.
\]
Thus, since $8\times60^{3/4}<195$, for sufficiently small $r$,
\[
    \E[|Y-\E[Y]|^3]
    \leq195w^3(1-w)^3r^6.
\]
\end{proof}

\subsection{Proof of Theorem \ref{thm:sample complexity 2-component}}
\begin{proof}
Let $P_r$, $Q_r$, $p_r$, and $q_r$ be as in Proposition \ref{prop:2-component analysis}. Let $X_1,\ldots,X_n\stackrel{\mathrm{i.i.d.}}{\sim}P_r$. Define $Y_i=\log p_r(X_i)-\log q_r(X_i)$ and $\hat{Y}=\frac{1}{n}\sum_{i=1}^nY_i$. Then
\[
    \mathbb{P}\left(\frac{1}{n}\sum_{i=1}^n\log p_r(X_i)\leq\frac{1}{n}\sum_{i=1}^n\log q_r(X_i)\right)=\mathbb{P}(\hat{Y}\leq0).
\]
Let $S_n = \sum_{i=1}^n Y_i = n\hat{Y}$. We wish to bound $\mathbb{P}(\hat{Y} \leq 0) = \mathbb{P}(S_n \leq 0)$.
We define the standardized random variable $Z_n$:
\begin{equation}
    Z_n = \frac{S_n - n\mu_Y}{\sqrt{n}\sigma_Y},
\end{equation}
where $\mu_Y = \mathbb{E}[Y]$ and $\sigma_Y = \sqrt{\text{Var}(Y)}$.

The event $\hat{Y} \leq 0$ corresponds to:
\begin{equation}
    Z_n \leq \frac{0 - n\mu_Y}{\sqrt{n}\sigma_Y} = -\sqrt{n} \frac{\mu_Y}{\sigma_Y}.
\end{equation}

The Berry-Esseen theorem implies the lower bound:
\begin{equation}
    \mathbb{P}\left(Z_n \leq x\right) \geq \Phi(x) - \frac{C \rho}{\sigma_Y^3 \sqrt{n}}.
\end{equation}
where $C$ is a universal constant ($C < 0.4748$) and $\rho = \E\left[|Y-\E[Y]|^3\right]$. 
It follows that
\begin{equation}
    \mathbb{P}(\hat{Y} \leq 0) \geq \Phi(-\sqrt{n} \frac{\mu_Y}{\sigma_Y}) - \frac{C \rho}{\sigma_Y^3 \sqrt{n}}.
\end{equation}

Let $\Gamma=[w(1-w)]^2$. From Proposition \ref{prop:2-component analysis}, we have
\begin{align}
    \mu_Y &= 4 r^4 \Gamma + O(r^6), \\
    \sigma_Y^2 &= 8 r^4 \Gamma + O(r^6), \\
    \rho &\leq 195r^6\Gamma^{3/2}.
\end{align}
For $r \ll 1$,
\begin{equation}
    \frac{\mu_Y}{\sigma_Y}
    =\sqrt{2}r^2\sqrt{\Gamma}\left(1+O(r^2)\right)
    <2r^2\sqrt{\Gamma}.
\end{equation}
Moreover, since $C<0.4748$,
\begin{equation}
    \frac{C\rho}{\sigma_Y^3}
    \leq \frac{195C}{8^{3/2}}\left(1+O(r^2)\right)<5
\end{equation}
for sufficiently small $r$. Therefore,
\begin{equation}
\mathbb{P}(\hat{Y} \leq 0)
\geq \Phi\left(-2\sqrt{n}r^2\sqrt{\Gamma}\right)-\frac{5}{\sqrt{n}}.
\end{equation}
Let $C_w=[2w(1-w)]^{-1/2}$. Since $n>25/\delta^2$, we have $5/\sqrt{n}<\delta$. Moreover, the condition on $r$ in the theorem gives
\[
    2\sqrt{n}r^2\sqrt{\Gamma}<-\Phi^{-1}(2\delta).
\]
Since $0<\delta<1/4$, it follows that
\[
    \Phi\left(-2\sqrt{n}r^2\sqrt{\Gamma}\right)>2\delta.
\]
It follows that
\[
    \mathbb{P}(\hat{Y}\leq0)>2\delta-\delta=\delta.
\]
This completes the proof of the theorem.
\end{proof}

\subsection{Proof of Theorem \ref{thm:minimax-testing}}
\label{app:minimax-proof}

We use Le Cam's two-point identity \cite{le2012asymptotic} and Pinsker's
inequality \cite[Lemma~2.5]{tsybakov2009introduction}, together with the
standard additivity of the Kullback--Leibler divergence for product measures.
For two probability measures $P,Q$ with product laws
$P^{\otimes n},Q^{\otimes n}$:

\begin{lemma}[Le Cam (two-point) lemma]
\label{lem:lecam-identity}
    $\displaystyle
    \inf_{\psi}\bigl[\mathbb E_{Q^{\otimes n}}\psi+\mathbb E_{P^{\otimes n}}(1-\psi)\bigr]
    =1-\mathrm{TV}\bigl(P^{\otimes n},Q^{\otimes n}\bigr).$
\end{lemma}

\begin{lemma}[Information bounds and tensorization]
\label{lem:tv-kl}
    $$
    \mathrm{TV}\bigl(P^{\otimes n},Q^{\otimes n}\bigr)
    \le\sqrt{\tfrac12\,D_{\mathrm{KL}}\!\bigl(P^{\otimes n}\,\|\,Q^{\otimes n}\bigr)}
    =\sqrt{\tfrac{n}{2}\,D_{\mathrm{KL}}(P\,\|\,Q)}.$$
\end{lemma}

\begin{proof}[Proof of Theorem~\ref{thm:minimax-testing}]
Proposition~\ref{prop:2-component analysis} gives, with
$g(z)=2w(1-w)(z^2-1)$ and $\E[g^2(Z)]=8w^2(1-w)^2$,
\begin{align}
    \label{eqn:kl-fixedvar}
    D_{\mathrm{KL}}(P_r\,\|\,Q_r)
    &=\E\!\left[\log\frac{p_r}{q_r}\right]
    =\tfrac12\,\E[g^2(Z)]\,r^4+O(r^6)\notag\\
    &=\kappa_w\,r^4+O(r^6),
    \qquad \kappa_w:=4w^2(1-w)^2 .
\end{align}
For any fixed $\varepsilon\in(0,1)$, there exists $r_0>0$ such that
\[
    D_{\mathrm{KL}}(P_r\|Q_r)
    \leq(1+\varepsilon)\kappa_w r^4,
    \qquad 0<r\leq r_0.
\]
Set $\Delta_0=2\sigma r_0$.

    Restricting the two suprema in \eqref{eqn:minimax-risk} to the single pair
    $P_r\in\mathcal P_\Delta$ and $Q_r\in\mathcal Q$ can only decrease the objective, so
    \[
        \mathcal R_n(\Delta)
        \;\ge\;
        \inf_\psi\bigl[\mathbb E_{Q_r^{\otimes n}}\psi+\mathbb E_{P_r^{\otimes n}}(1-\psi)\bigr]
        \;=\;
        1-\mathrm{TV}\bigl(P_r^{\otimes n},Q_r^{\otimes n}\bigr)
    \]
    by Lemma~\ref{lem:lecam-identity}. Applying Lemma~\ref{lem:tv-kl} and then
    \eqref{eqn:kl-fixedvar},
    \[
        \mathrm{TV}\bigl(P_r^{\otimes n},Q_r^{\otimes n}\bigr)
        \le\sqrt{\tfrac{n}{2}(1+\varepsilon)\kappa_w r^4}
        =\sqrt{1+\varepsilon}\,
        \frac{w(1-w)}{2\sqrt2}\,\frac{\Delta^2}{\sigma^2}\sqrt n,
    \]
    using $\kappa_w=4w^2(1-w)^2$ and $r=\Delta/(2\sigma)$. This is
    \eqref{eqn:minimax-lb}. Solving
    \[
        \sqrt{1+\varepsilon}\,\frac{w(1-w)}{2\sqrt2}
        \frac{\Delta^2}{\sigma^2}\sqrt n\le 1-\eta
    \]
    for $\Delta$ and
    setting $\eta=\tfrac12$ gives the two displayed thresholds.
\end{proof}

\subsection{Proof of Proposition \ref{prop:k-component}}
\begin{proof}
    By the generating function of the probabilists' Hermite polynomials,
    \[
        \phi(x-t)=\phi(x)\sum_{j=0}^{\infty}\frac{H_j(x)}{j!}t^j.
    \]
    Therefore,
    \[
        p_R(x)=\phi(x)\sum_{j=0}^{\infty}
        \frac{H_j(x)}{j!}m_j(\overline{\nu})R^j,
    \]
    while
    \[
        q_R(x)=\phi(x)\sum_{j=0}^{\infty}
        \frac{H_j(x)}{j!}m_j(\overline{\nu}')R^j.
    \]
    Since the zeroth moments agree and the first $2k-3$ moments are matched,
    \[
        p_R(x)-q_R(x)
        =\phi(x)g(x)R^{2k-2}+o(R^{2k-2}).
    \]
    Moreover, $q_R(x)\to\phi(x)$ as $R\to0$. Hence,
    \[
        S_R(x):=\frac{p_R(x)}{q_R(x)}
        =1+g(x)R^{2k-2}+o(R^{2k-2}).
    \]
    The remainder in this expansion is sufficiently controlled when taking moments up to order three. Also, $\E[g(Z)]=0$ because $\E[H_{2k-2}(Z)]=0$ for $Z\sim\mathcal{N}(0,1)$.

    Let $u_R=S_R-1$. Since $p_R$ and $q_R$ are probability densities,
    \[
        \int_{\mathbb{R}}q_R(x)u_R(x)\,dx
        =\int_{\mathbb{R}}\bigl(p_R(x)-q_R(x)\bigr)\,dx=0.
    \]
    By the change-of-measure formula,
    \begin{align*}
        \E[Y]
        &=\int_{\mathbb{R}}p_R(x)\log\frac{p_R(x)}{q_R(x)}\,dx\\
        &=\int_{\mathbb{R}}q_R(x)(1+u_R(x))\log(1+u_R(x))\,dx.
    \end{align*}
    Using
    \[
        (1+u)\log(1+u)=u+\frac{u^2}{2}+O(|u|^3),
    \]
    together with the vanishing of the linear term, we obtain
    \[
        \E[Y]=\frac{1}{2}\E[g^2(Z)]R^{4k-4}+o(R^{4k-4}).
    \]

    Similarly,
    \begin{align*}
        \E[Y^2]
        &=\int_{\mathbb{R}}q_R(x)(1+u_R(x))
        \bigl(\log(1+u_R(x))\bigr)^2\,dx\\
        &=\E[g^2(Z)]R^{4k-4}+o(R^{4k-4}).
    \end{align*}
    Since
    \[
        \bigl(\E[Y]\bigr)^2=O(R^{8k-8})=o(R^{4k-4}),
    \]
    it follows that
    \[
        \operatorname{Var}(Y)
        =\E[g^2(Z)]R^{4k-4}+o(R^{4k-4}).
    \]

    Finally, using $\log(1+u_R)=u_R+O(u_R^2)$ gives
    \[
        Y=g(X)R^{2k-2}+o(R^{2k-2}).
    \]
    Since $\E[Y]=O(R^{4k-4})=o(R^{2k-2})$,
    \[
        Y-\E[Y]=g(X)R^{2k-2}+o(R^{2k-2}).
    \]
    As $R\to0$, $P_R$ converges to $\mathcal{N}(0,1)$, and the corresponding absolute moments converge. Therefore,
    \[
        \E\left[|Y-\E[Y]|^3\right]
        =\E[|g(Z)|^3]R^{6k-6}+o(R^{6k-6}).
    \]
    This completes the proof.
\end{proof}

\subsection{Proof of Theorem \ref{thm:sample complexity of k-component}}
\begin{proof}
    Let
    \[
        A_{k,w}=\E[g^2(Z)],
        \qquad
        B_{k,w}=\E[|g(Z)|^3],
        \qquad Z\sim\mathcal{N}(0,1).
    \]
    We first note that $A_{k,w}>0$. Indeed, let
    \[
        \ell(x)=\prod_{i=1}^{k-1}(x-b_i).
    \]
    Since $\overline{\nu}'$ is supported on $b_1,\ldots,b_{k-1}$,
    \[
        \int \ell^2(x)\,d\overline{\nu}'(x)=0,
    \]
    whereas $\int \ell^2(x)\,d\overline{\nu}(x)>0$ because $\overline{\nu}$ has $k$ distinct support points with positive weights. Since $\ell^2$ is monic of degree $2k-2$ and the first $2k-3$ moments are matched,
    \[
        m_{2k-2}(\overline{\nu})-m_{2k-2}(\overline{\nu}')>0,
    \]
    and hence $g$ is not identically zero.

    Let $X_1,\ldots,X_n\stackrel{\mathrm{i.i.d.}}{\sim}P_R$ and define
    \[
        Y_i=\log\frac{p_R(X_i)}{q_R(X_i)},
        \qquad
        \overline{Y}_n=\frac{1}{n}\sum_{i=1}^nY_i.
    \]
    Write
    \[
        \mu_R=\E[Y_i],
        \qquad
        \sigma_R^2=\operatorname{Var}(Y_i),
        \qquad
        \rho_R=\E[|Y_i-\mu_R|^3].
    \]
    By Proposition \ref{prop:k-component},
    \begin{align*}
        \mu_R&=\frac{1}{2}A_{k,w}R^{4k-4}+o(R^{4k-4}),\\
        \sigma_R^2&=A_{k,w}R^{4k-4}+o(R^{4k-4}),\\
        \rho_R&=B_{k,w}R^{6k-6}+o(R^{6k-6}).
    \end{align*}
    Consequently,
    \[
        \frac{\mu_R}{\sigma_R}
        =\frac{1}{2}\sqrt{A_{k,w}}R^{2k-2}(1+o(1)),
        \qquad
        \frac{C\rho_R}{\sigma_R^3}
        =\frac{CB_{k,w}}{A_{k,w}^{3/2}}+o(1),
    \]
    where $C$ is the universal Berry--Esseen constant. Thus, there exist $R_0>0$ and $M_{k,w}>0$ such that, for $0<R<R_0$,
    \[
        \frac{\mu_R}{\sigma_R}\leq\sqrt{A_{k,w}}R^{2k-2},
        \qquad
        \frac{C\rho_R}{\sigma_R^3}\leq M_{k,w}.
    \]

    Define
    \[
        N_{k,w,\delta}
        =\left\lceil\left(\frac{M_{k,w}}{\delta}\right)^2\right\rceil,
        \qquad
        C_{k,w}=A_{k,w}^{-\frac{1}{4k-4}}.
    \]
    By the Berry--Esseen theorem,
    \[
        \mathbb{P}(\overline{Y}_n\leq0)
        \geq
        \Phi\left(-\sqrt{n}\frac{\mu_R}{\sigma_R}\right)
        -\frac{C\rho_R}{\sigma_R^3\sqrt{n}}.
    \]
    If $n\geq N_{k,w,\delta}$, the second term is at most $\delta$. Moreover, the condition on $R$ in the theorem gives
    \[
        \sqrt{n}\frac{\mu_R}{\sigma_R}
        \leq\sqrt{nA_{k,w}}R^{2k-2}
        <-\Phi^{-1}(2\delta).
    \]
    Since $0<\delta<1/4$, it follows that
    \[
        \Phi\left(-\sqrt{n}\frac{\mu_R}{\sigma_R}\right)>2\delta.
    \]
    Therefore,
    \[
        \mathbb{P}(\overline{Y}_n\leq0)>2\delta-\delta=\delta,
    \]
    which proves the theorem.
\end{proof}

\subsection{Proof of Theorem \ref{thm:minimax-k}}
\begin{proof} Following the notations in 
Proposition~\ref{prop:k-component}, we 
write
\[
    g(x)=\frac{m_{2k-2}(\overline\nu)-m_{2k-2}(\overline\nu')}{(2k-2)!}\,H_{2k-2}(x),
\]
the Hermite orthogonality relation $\mathbb E[H_j(Z)^2]=j!$ gives the closed form
\begin{equation}
    \label{eqn:Eg2}
    \mathbb E[g^2(Z)]
    =\frac{\bigl(m_{2k-2}(\overline\nu)-m_{2k-2}(\overline\nu')\bigr)^2}{(2k-2)!}
    \;=:\;\tau_{k,w}^2 .
\end{equation}

Proposition~\ref{prop:k-component} then yields
\begin{equation}
    \label{eqn:kl-k}
    D_{\mathrm{KL}}(P_R\,\|\,Q_R)
    =\mathbb E[Y]
    =\tfrac12\,\mathbb E[g^2(Z)]\,R^{4k-4}+o(R^{4k-4})
    =\kappa_{k,w}\,\Delta^{4k-4}+o(\Delta^{4k-4}),
\end{equation}
where, using $R=\tfrac{k-1}{2}\Delta$,
\[
    \kappa_{k,w}
    :=\tfrac12\,\tau_{k,w}^2\Bigl(\tfrac{k-1}{2}\Bigr)^{4k-4}.
\]
For any fixed $\varepsilon\in(0,1)$, \eqref{eqn:kl-k} implies that there exists
$\Delta_0>0$ such that
\[
    D_{\mathrm{KL}}(P_R\|Q_R)
    \leq(1+\varepsilon)\kappa_{k,w}\Delta^{4k-4},
    \qquad 0<\Delta\leq\Delta_0.
\]

    Restrict the two suprema in \eqref{eqn:minimax-risk-k} to the single pair
    $P_R\in\mathcal P_\Delta^{(k)}$ and $Q_R\in\mathcal Q^{(k-1)}$; this only decreases the
    objective, so by Lemma~\ref{lem:lecam-identity},
    $\mathcal R_n^{(k)}(\Delta)\ge 1-\mathrm{TV}(P_R^{\otimes n},Q_R^{\otimes n})$.
    Lemma~\ref{lem:tv-kl} and \eqref{eqn:kl-k} give
    $\mathrm{TV}(P_R^{\otimes n},Q_R^{\otimes n})
    \le\sqrt{\tfrac n2(1+\varepsilon)\kappa_{k,w}\Delta^{4k-4}}$, which is
    \eqref{eqn:minimax-lb-k}. Solving $\sqrt{\tfrac{\kappa_{k,w}}{2}}\Delta^{2k-2}\sqrt n\le1-\eta$
    with $\kappa_{k,w}$ replaced by $(1+\varepsilon)\kappa_{k,w}$
    for $n$ and $\Delta$ gives the two thresholds; $\eta=\tfrac12$ gives the last inequality.
\end{proof}

\subsection{Proof of Proposition \ref{prop:concentration}}
\begin{proof}
    Notice that
    \[
        \E[e^{\iota tX_j}]
        =\exp\left(-\frac{\sigma^2t^2}{2}\right)\mathcal{F}[\nu](t),
    \]
    and hence $\E[y_n(t)]=\mathcal{F}[\nu](t)$. Applying Hoeffding's inequality to the real and imaginary parts of $y_n(t)-\mathcal{F}[\nu](t)$, we have that for any $u>0$,
    \begin{align*}
        \mathbb{P}\left(\left|\operatorname{Re}\{y_n(t)-\mathcal{F}[\nu](t)\}\right|\geq u\right)
        &\leq 2\exp\left(-\frac{nu^2}{2\exp(\sigma^2t^2)}\right),\\
        \mathbb{P}\left(\left|\operatorname{Im}\{y_n(t)-\mathcal{F}[\nu](t)\}\right|\geq u\right)
        &\leq 2\exp\left(-\frac{nu^2}{2\exp(\sigma^2t^2)}\right).
    \end{align*}
    Here, the real and imaginary summands before centering lie in
    $[-\exp(\sigma^2t^2/2),\exp(\sigma^2t^2/2)]$.
    Therefore,
    \begin{align*}
        \mathbb{P}\left(\left|y_n(t)-\mathcal{F}[\nu](t)\right|\geq\epsilon\right)
        &\leq \mathbb{P}\left(\left|\operatorname{Re}\{y_n(t)-\mathcal{F}[\nu](t)\}\right|\geq\frac{\epsilon}{\sqrt{2}}\right)\\
        &\quad+\mathbb{P}\left(\left|\operatorname{Im}\{y_n(t)-\mathcal{F}[\nu](t)\}\right|\geq\frac{\epsilon}{\sqrt{2}}\right)\\
        &\leq 4\exp\left(-\frac{n\epsilon^2}{4\exp(\sigma^2t^2)}\right).
    \end{align*}
\end{proof}

\subsection{Proof of Theorem \ref{thm:thresholding}}
Before proving Theorem \ref{thm:thresholding}, we introduce some useful notation and preliminary results. We define the noiseless Hankel matrix and noise matrix:
\begin{equation*}
        Y = \begin{bmatrix}
            y_0 & y_1 & \cdots & y_L   \\
            y_1 & y_2 & \cdots & y_{L+1}\\
            \vdots     & \vdots     & \ddots & \vdots        \\
            y_L & y_{L+1}&\cdots & y_{2L}
        \end{bmatrix} \in \C^{(L+1)\times (L+1)}, \quad \Lambda = H - Y,
    \end{equation*}
where $y_q = \sum_{i=1}^k w_i \exp(\iota \mu_i t_q)$ for $q = 0,1, \cdots, 2L$. We introduce the following Vandermonde vector
    \[
        \phi_{L}(\mu) = \begin{bmatrix}
            1 & \exp\left(\iota \frac{f}{L}\mu\right) & \exp\left(\iota 2\frac{f}{L}\mu\right) & \cdots & \exp\left(\iota L\frac{f}{L}\mu\right)
        \end{bmatrix}^\mathrm{T} \in \C^{L+1}.
    \]
The matrix $Y$ admits the Vandermonde decomposition:
    \[
        Y = \sum_{i=1}^k w_i \exp\left(-\iota f \mu_i\right) \phi_L(\mu_i) \phi_L(\mu_i)^\mathrm{T} = \Phi_L D \Phi_L^\mathrm{T},
    \]
where $\Phi_L = \left[\phi_L(\mu_1)\ \cdots\ \phi_L(\mu_k)\right] \in \C^{(L+1)\times k}$ and $D = \text{diag}\left(w_1 e^{-\iota f \mu_1}, \cdots, w_k e^{-\iota f \mu_k}\right) \in \C^{k\times k}$. It follows that the rank of the matrix $Y$ is $k$, which is exactly the order of the mixture. The following Weyl's theorem is needed to estimate the perturbation of the singular values:
\begin{proposition}(\cite{weyl1912asymptotische})
\label{Wely}
Let $M$ be a $m\times n$ matrix, and $s_l(M)$ its $l$-th singular value. Let $\Lambda \in \mathbb{C}^{m\times n}$ be a perturbation to $M$. 
Then the following bound holds for the perturbed singular values
    $$
        |s_l(M+ \Lambda) - s_l(M)| \leq ||\Lambda||_2.  
    $$
\end{proposition}

The following proposition and lemma are needed to complete the proof.

\begin{proposition} (Theorem 1 in \cite{gautschi1962inverses}).
\label{prop:van}
Let $x_i \neq x_j$ for $i\neq j$. 
Let 
    \[
        V_k = \begin{bmatrix}
        		1&1&\cdots&1 \\
                    x_1&x_2&\cdots&x_k \\
                    \vdots&\vdots&\ddots&\vdots\\
                    x_1^{k-1}&x_2^{k-1}&\cdots&x_k^{k-1}
        	\end{bmatrix} 
    \]
be a Vandermonde matrix. 
Then
    \begin{equation}
    \label{eqn:2normVan}
        ||V_k^{-1}||_\infty \leq \max_{1\leq j\leq k} \prod_{i=1,i\neq j}^k \frac{1+|x_i|}{|x_i-x_j|}.
    \end{equation}
\end{proposition}

\begin{lemma}\label{lem: smallest singular value}
    Let $L > k-1$, notice
    \begin{align*}
        & \Phi_{k-1} = \left[\phi_{k-1}(\mu_1),\phi_{k-1}(\mu_2),\cdots,\phi_{k-1}(\mu_k)\right] \in \mathbb{C}^{k\times k}, \\
        & \Phi_{L} = \left[\phi_{L}(\mu_1),\phi_{L}(\mu_2),\cdots,\phi_{L}(\mu_k)\right] \in \mathbb{C}^{(L+1)\times k}.
    \end{align*}
   Then
    \begin{equation*}
        s_*(\Phi_{L}) \geq s_*(\Phi_{k-1}) \geq \frac{1}{\norm{\Phi_{k-1}^{-1}}}.
    \end{equation*}
    Here $s_*(A)$ denotes the smallest singular value of a matrix $A$.
\end{lemma}
The proof of the lemma can be found in Proposition 4.4 in \cite{liu2021theory}. The following lemma is used to estimate the smallest nonzero singular value of $Y$:

\begin{lemma}
\label{lem:minimal singular value}
    Suppose that $k \geq 2$ and $\mu_i \in \left[-\frac{(k-1)\pi}{2f}, \frac{(k-1)\pi}{2f}\right]$. Then the following estimation holds:
    \[
        s_k(Y) \geq \frac{w_{\min}\zeta(k)^2 (f\Delta)^{2k-2}}{k(\pi L)^{2k-2}}
    \]
\end{lemma}

\begin{proof}
    Denote $\text{ker}(\Phi^\mathrm{T})$ as the kernel space of $\Phi^\mathrm{T}$ and $\text{ker}^\bot(\Phi^\mathrm{T})$ as its orthogonal complement. We have
    \begin{equation*}
        s_k(Y) = s_*(\Phi_L D \Phi_L^\mathrm{T}) = \min_{\substack{\norm{x}_2=1\\ x \in \ker^{\bot}(\Phi_L^{\mathrm{T}})}}\norm{\Phi_L D \Phi_L^\mathrm{T} x} \geq s_*(\Phi_L D)s_*(\Phi_L^{\mathrm{T}}) \geq s_*(\Phi_L) s_*(D)s_*(\Phi_L).
    \end{equation*}
    After relabeling the components, assume that $\mu_1<\cdots<\mu_k$. Since $L>k-1$, by Proposition \ref{prop:van} and Lemma \ref{lem: smallest singular value}, we have
    \begin{align*}
        s_*(\Phi_L) \geq s_*(\Phi_{k-1}) \geq \frac{1}{\norm{\Phi_{k-1}^{-1}}} & \geq \frac{1}{\sqrt{k}\norm{\Phi_{k-1}^{-1}}_\infty}\\
        &\geq \frac{1}{\sqrt{k}} \min_{1\leq j \leq k} \prod_{i=1,i\neq j}^k \frac{|\exp\left(\iota \frac{f}{L}\mu_i\right) - \exp\left(\iota \frac{f}{L}\mu_j\right)|}{2}.
    \end{align*}
    Notice that $\mu_i \in [-\frac{(k-1)\pi}{2f}, \frac{(k-1)\pi}{2f}]$ for $i =1,\cdots,k$, we have
    \[
        \left|\exp\left(\iota \frac{f}{L}\mu_i\right) - \exp\left(\iota \frac{f}{L}\mu_j\right)\right| \geq \frac{2f}{\pi L}\abs{\mu_i - \mu_j}.
    \]
    It follows that
    \begin{align*}
        \prod_{i=1,i\neq j}^k \frac{|\exp\left(\iota \frac{f}{L}\mu_i\right) - \exp\left(\iota \frac{f}{L}\mu_j\right)|}{2}
        &\geq (\frac{f}{\pi L})^{k-1} \prod_{i=1,i\neq j}^k \abs{\mu_i - \mu_j} \\
        &\geq (\frac{f}{\pi L})^{k-1} \prod_{i < j}^k \abs{\mu_i - \mu_j} \prod_{i > j}^k \abs{\mu_i - \mu_j} \\
        &\geq (\frac{f}{\pi L})^{k-1} (j-1)! \Delta^{j-1} (k-j)! \Delta^{k-j} \\
        &= (\frac{\Delta f}{\pi L})^{k-1} (j-1)!(k-j)!.
    \end{align*}
    Thus, we have
    \[
        s_*(\Phi_L) \geq \frac{\zeta(k)}{\sqrt{k}}(\frac{\Delta f}{\pi L})^{k-1}.
    \]
    Furthermore,
    \begin{align*}
        s_k(Y) \geq \frac{\zeta(k)^2}{k}(\frac{\Delta f}{\pi L})^{2k-2} s_*(D) \geq \frac{\zeta(k)^2}{k}(\frac{\Delta f}{\pi L})^{2k-2} w_{\min}.
    \end{align*}
\end{proof}

We next present the proof of the main theorem:
\begin{proof}
    Define the maximum-entry norm of a matrix by $\norm{A}_{\max}=\max_{a,b}|A_{ab}|$. Notice that
    \[
        \norm{\Lambda}_{\max}=\max_{0\leq q\leq 2L}
        |y_n(t_q)-\mathcal{F}[\nu](t_q)|.
    \]
    Applying Proposition \ref{prop:concentration} and using $|t_q|\leq f$, we obtain
    \begin{align*}
        \mathbb{P}\left(\norm{\Lambda}_{\max}\geq\epsilon\right)
        &\leq\sum_{q=0}^{2L}\mathbb{P}\left(
        |y_n(t_q)-\mathcal{F}[\nu](t_q)|\geq\epsilon\right)\\
        &\leq\sum_{q=0}^{2L}4\exp\left(-\frac{n\epsilon^2}
        {4\exp(\sigma^2t_q^2)}\right)\\
        &\leq4(2L+1)\exp\left(-\frac{n\epsilon^2}
        {4\exp(\sigma^2f^2)}\right).
    \end{align*}
    If the sample size $n$ satisfies \eqref{eqn:thresholding sample size}, we have 
    \[
        \mathbb{P}\left(\norm{\Lambda}_{\max}<\epsilon\right)\geq1-\delta.
    \]
Thus, with probability at least $1-\delta$, we have $\norm{\Lambda}_2 \leq \norm{\Lambda}_F < (L+1)\epsilon$. By applying Proposition \ref{Wely}, we have 
\[
    |s_i(H) - s_i(Y)| \leq \norm{\Lambda}_2 < (L+1)\epsilon,\quad i = 1,\cdots,(L+1).
\]
Since $s_{i}(Y) = 0$ for $i = k+1,\cdots,L+1$, we have
    \[
        s_i(H) < (L+1) \epsilon
    \]
for $i = k+1, \cdots , L+1$. Moreover, by \eqref{eqn:thresholding epsilon} and Lemma \ref{lem:minimal singular value},
    \begin{align*}
        s_k(H)
        &\geq s_k(Y)-\norm{\Lambda}_2\\
        &>2(L+1)\epsilon-(L+1)\epsilon
        =(L+1)\epsilon,
    \end{align*}
which proves the theorem.
\end{proof}

\subsection{Proofs for Theorem~\ref{thm:thresholding-single} and Corollary~\ref{cor:sampling-single} }
\label{app:local-rates}

Recall the following notation:
$\Phi_L\in\mathbb{C}^{(L+1)\times k}$ is the truncated Fourier matrix with entries
$[\Phi_L]_{r,i}=e^{\iota r\frac{f}{L}\mu_i}$, $D=\operatorname{diag}(w_1,\dots,w_k)$ is the
diagonal weight matrix, $Y=\Phi_L D\Phi_L^{\mathrm T}$ is the noiseless Hankel
matrix, and $\Lambda$ is the empirical perturbation.

We first present some basic geometric estimates. The rows of
$\Phi_L$ evaluate the monomials $z_i^{\,r}$ at the unit-modulus nodes
\[
    z_i \;=\; e^{\iota\omega_i}, \qquad \omega_i \;=\; \tfrac{f}{L}\,\mu_i,
    \qquad i=1,\dots,k,
\]
so that 
\begin{equation}
    \label{eqn:chord}
    \frac{|z_i-z_j|}{2}
    \;=\; \left|\sin\!\left(\frac{\omega_i-\omega_j}{2}\right)\right|
    \;=\; \left|\sin\!\left(\frac{f\,(\mu_i-\mu_j)}{2L}\right)\right|.
\end{equation}
Under the hypothesis $L\ge k$, for
$\mu_i\in[-\frac{(k-1)\pi}{2f},\frac{(k-1)\pi}{2f}]$, every gap satisfies
$|\mu_i-\mu_j|\le\frac{(k-1)\pi}{f}$, whence
\begin{equation}
    \label{eqn:angle-bound}
    \frac{f\,|\mu_i-\mu_j|}{2L}
    \;\le\; \frac{(k-1)\pi}{2L}
    \;\le\; \frac{(k-1)\pi}{2k}
    \;<\; \frac{\pi}{2}.
\end{equation}
Thus \emph{all} angular gaps lie in $[0,\tfrac{\pi}{2})$, where the sine is
positive, strictly increasing, and obeys the linear lower bound
$\sin\theta\ge\frac{2}{\pi}\theta$.

Under the single-collision configuration of Definition~\ref{def:single-collision}, for $j\in\{1,2\}$ the merged partner is excluded,
so all $k-2$ surviving factors involve resolved separations
$f|\mu_i-\mu_j|\in[c_0,(k-1)\pi]$; applying the lower bound
$\sin\theta\ge\frac{2}{\pi}\theta$ to each factor gives
$\sin\!\left(\frac{f|\mu_i-\mu_j|}{2L}\right)\ge\frac{c_0}{\pi L}$, and hence 
\begin{equation}
    \label{eqn:resolved-product}
    \prod_{\substack{i=1\\ i\neq j,\ \{i,j\}\neq\{1,2\}}}^{k}
    \sin\!\left(\frac{f\,|\mu_i-\mu_j|}{2L}\right)
    \ \ge\
    \left(\frac{c_0}{\pi L}\right)^{k-2},
    \qquad j\in\{1,2\}.
\end{equation}
This leads to the following Singular value estimate for the single-collision configuration. 

\begin{lemma}
\label{lem:sigmak-single}
    Suppose that $L \geq k$ and $\mu_i \in \left[-\frac{(k-1)\pi}{2f},
    \frac{(k-1)\pi}{2f}\right]$ for $i=1,\ldots,k$, and that the nodes satisfy the
    single-collision configuration of Definition~\ref{def:single-collision}. Then the
    noiseless Hankel matrix $Y=\Phi_L D\Phi_L^{\mathrm T}$ satisfies
    \[
        s_k(Y)\ \geq\ \frac{w_{\min}\,c_0^{2(k-2)}\,(f\Delta)^{2}}
        {k\,(\pi L)^{2k-2}}.
    \]
\end{lemma}

\begin{proof}
    By the factorization $Y=\Phi_L D\Phi_L^{\mathrm T}$ and Weyl's inequality,
    $s_k(Y)\ge w_{\min}\,s_*(\Phi_L)^2$. Since $L\ge k$, restricting to the
    top $k$ rows gives $s_*(\Phi_L)\ge s_*(V)$ for the square Vandermonde
    $V=[z_i^{\,r}]_{0\le r\le k-1,\,1\le i\le k}$. Lemma~\ref{lem: smallest singular value} yields
    \[
        s_*(V)\ \ge\ \frac{1}{\sqrt{k}}\,
        \min_{1\le j\le k}\ \prod_{i\neq j}\frac{|z_i-z_j|}{2}
        \ =\ \frac{1}{\sqrt{k}}\,
        \min_{1\le j\le k}\ \prod_{i\neq j}\sin\!\left(\frac{f\,|\mu_i-\mu_j|}{2L}\right),
    \]
    where we used the identity~\eqref{eqn:chord} and the fact that all angles
    lie in $[0,\tfrac{\pi}{2})$ by~\eqref{eqn:angle-bound}. For the merged pair we
    apply the linear bound $\sin\theta\ge\frac{2}{\pi}\theta$:
    \[
        \sin\!\left(\frac{f\Delta}{2L}\right)
        \ \ge\ \frac{2}{\pi}\cdot\frac{f\Delta}{2L}
        \ =\ \frac{f\Delta}{\pi L}.
    \]
    For an index $j\ge 3$ the product $\prod_{i\ne j}$ contains only resolved
    (order-one) factors, while for $j\in\{1,2\}$ it contains exactly one small factor
    $\frac{f\Delta}{\pi L}$ together with $k-2$ resolved factors. Since $f\Delta\ll1$,
    the minimum is attained at a merged node, and combining the small factor with the
    resolved-product bound~\eqref{eqn:resolved-product},
    \[
        \min_{1\le j\le k}\ \prod_{i\neq j}\frac{|z_i-z_j|}{2}
        \ \ge\ \frac{f\Delta}{\pi L}\left(\frac{c_0}{\pi L}\right)^{k-2}
        \ =\ \frac{c_0^{\,k-2}\,(f\Delta)}{(\pi L)^{k-1}}.
    \]
    Combining the displays gives
    $s_*(\Phi_L)\ge\frac{1}{\sqrt{k}}\,\frac{c_0^{\,k-2}(f\Delta)}{(\pi L)^{k-1}}$,
    and squaring and multiplying by $w_{\min}$ completes the proof.
\end{proof}

\begin{proof}[Proof of Theorem~\ref{thm:thresholding-single}]
    The first assertion is identical to Theorem~\ref{thm:thresholding} and does not
    use the configuration: under~\eqref{eqn:thresholding-single sample size}, a
    Bernstein bound gives $\|\Lambda\|\le(L+1)\epsilon$ with probability at least
    $1-\delta$; since $\operatorname{rank}Y=k$, Weyl's inequality yields
    $s_j=s_j(Y+\Lambda)\le s_j(Y)+\|\Lambda\|=\|\Lambda\|\le(L+1)\epsilon$
    for all $j\ge k+1$. For the signal side, Weyl gives
    $s_k\ge s_k(Y)-\|\Lambda\|\ge s_k(Y)-(L+1)\epsilon$. Inserting
    Lemma~\ref{lem:sigmak-single} and using~\eqref{eqn:thresholding-single epsilon},
    which enforces $(L+1)\epsilon<\tfrac12 s_k(Y)$, we obtain
    $s_k>s_k(Y)-\tfrac12 s_k(Y)=\tfrac12 s_k(Y)>(L+1)\epsilon$.
\end{proof}

\begin{proof}[Proof of Corollary~\ref{cor:sampling-single}]
    Substituting $\epsilon=\epsilon_*$ into~\eqref{eqn:thresholding-single sample size}
    and using
    $$\epsilon_*^2=\frac{w_{\min}^2 c_0^{4(k-2)}(f\Delta)^{4}}
    {16k^2(\pi L)^{4k-4}(L+1)^2}$$
    gives the stated sample size~\eqref{eqn:sampling-single}. Since
    $\epsilon_*=\tfrac12$ times the right-hand side
    of~\eqref{eqn:thresholding-single epsilon}, both the noise bound and the signal
    bound of Theorem~\ref{thm:thresholding-single} hold simultaneously, so the
    threshold $(L+1)\epsilon_*$ separates $s_k$ from $s_{k+1}$ and the
    algorithm outputs exactly $k$.
\end{proof}

\subsection{Hermite Polynomials and Moment-matching}
\label{subsec:moment-matching}
In the asymptotic analysis of a $k$-component GMM, we decompose its density using the Hermite polynomials. The ``probabilists' Hermite polynomials'' are defined by
    \[
        H_n(x) = (-1)^n e^{\frac{x^2}{2}}\frac{d^n}{dx^n}e^{-\frac{x^2}{2}},\quad n = 0, 1, \cdots, 
    \]
which is an orthogonal polynomial series. For a more comprehensive study of Hermite polynomials, see \cite{szeg1939orthogonal}. We shall use the following important property of the Hermite polynomials:
\begin{proposition}
\label{prop:generating function}
    (Generating function) The Hermite polynomials have the generating function:
    \[
    e^{xt - \frac{t^2}{2}} = \sum_{n=0}^{+\infty} H_n(x)\frac{t^n}{n!}.
    \]
    Then if $X \sim  \mathcal{N}(\mu, 1)$, we have that $\E[H_n(X)] = \mu^n$.
\end{proposition}
\begin{proof}
    The proof of the generating function can be found in \cite{szeg1939orthogonal}. Now if $X \sim \mathcal{N}(\mu, 1)$, then its moment generating function is 
    \[
        M_X(t) = \E[e^{tX}] = e^{\mu t + \frac{t^2}{2} }.
    \]
    We can compute that
    \[
        \E[e^{Xt - \frac{t^2}{2}}] = e^{-\frac{t^2}{2}} \E[e^{tX}] = e^{\mu t} = \sum_{n=0}^{+\infty}\frac{\mu^n}{n!}t^n.
    \]
    On the other hand, by the generating function, we have
    \[
        \E[e^{Xt - \frac{t^2}{2}}] = \E\left[\sum_{n=0}^{+\infty} H_n(X)\frac{t^n}{n!}\right] = \sum_{n=0}^{+\infty} \frac{\E[H_n(X)]}{n!} t^n.
    \]
    By equating the coefficients of the above two equations, we prove the statement.
\end{proof}

The asymptotic analysis of the divergence 
in Proposition \ref{prop:k-component} is based on the moment-matching. 
We define the $r$-th moment vector of distribution $\nu$ as
\[
    \vm_r = \bigl(m_1(\nu),m_2(\nu),\ldots,m_r(\nu)\bigr).
\]
The $r$-th moment space supported on $S \subset \R$ is defined as
    \[
        \mathcal{M}_r(S) = \{\vm_r(\nu): \nu \text{ is supported on } S\},
    \]
which is a convex set. For details, one can refer to \cite{shohat1943problem}.

The solvability of the moment-matching equation for distributions $\nu$ and $\nu'$ supported on a finite subset of $\R$:
    \begin{equation}
    \label{eqn:moment-matching}
        \vm_{2k-3}(\nu') = \vm_{2k-3}(\nu)
    \end{equation}
is based on the theory of Gaussian quadrature, which shows that any valid $(2k-3)$-th moment vector can be realized by a unique $(k-1)$-component discrete distribution $\nu'=\sum_{i=1}^{k-1} \pi_i \delta_{\phi_i}$. The computation of such a unique distribution is based on Algorithm \ref{algo:moment-matching} below. This algorithm relies on the root-finding of high degree polynomials, which may suffer from instability issues. For the improvements of the algorithm, see \cite{gautschi2004orthogonal}.


    \begin{algorithm}
    \label{algo:moment-matching}
        \caption{Gaussian Quadrature}
        \Input{moment vector $\vm_{2k-3} = (m_1, \cdots, m_{2k-3})$}
        $\phi_1, \cdots, \phi_{k-1} \gets $ roots of polynomial
        \[
            \Pi_{k-1}(x) = \det \begin{bmatrix}
                1 & m_1 & \cdots & m_{k-1} \\
                \vdots & \vdots & \ddots & \vdots \\
                m_{k-2} & m_{k-1} & \cdots & m_{2k-3} \\
                1 & x & \cdots & x^{k-1}
            \end{bmatrix};
        \]

        Let $\vw = (\pi_1, \cdots, \pi_{k-1})^\mathrm{T} $ and 
        \[
            \vw \gets \begin{bmatrix}
                1 & 1 & \cdots & 1 \\
                \phi_1 & \phi_2 & \cdots & \phi_{k-1} \\
                \vdots & \vdots & \ddots & \vdots \\
                \phi_1^{k-2} & \phi_2^{k-2} & \cdots & \phi_{k-1}^{k-2}
            \end{bmatrix}^{-1} \begin{bmatrix}
                1 \\ m_1 \\ \vdots \\ m_{2k-3}
            \end{bmatrix};
        \]
        
    \Output{discrete distribution $\sum_{i=1}^{k-1} \pi_i \delta_{\phi_i}$.}
    \end{algorithm}

\bibliographystyle{plainnat}
\bibliography{reference}

\end{document}